\documentclass{article}
\usepackage{ijcai25}

\usepackage{times}
\usepackage{soul}
\usepackage{url}
\usepackage[hidelinks]{hyperref}
\usepackage[utf8]{inputenc}
\usepackage[small]{caption}
\usepackage{graphicx}
\usepackage{amsmath}
\usepackage{amsthm}
\usepackage{booktabs}
\usepackage{algorithm}
\usepackage{algorithmic}
\usepackage{makecell} 
\usepackage[switch]{lineno}
\usepackage{tikz}
\usetikzlibrary{positioning}
\usepackage{xcolor}

\title{A Survey on Human-Centered Evaluation of Explainable AI Methods in Clinical Decision Support Systems}

\author{
Alessandro Gambetti $^{1,2}$
\and
Qiwei Han$^2$\and
Hong Shen$^3$\And
Cl\'audia Soares$^1$\\
\affiliations
$^1$Nova School of Science and Technology, Universidade NOVA de Lisboa, Caparica, Portugal\\
$^2$Nova School of Business and Economics, Universidade NOVA de Lisboa, Carcavelos, Portugal \\
$^3$Carnegie Mellon University, Pittsburgh PA \\
\emails
a.gambetti@campus.fct.unl.pt,
qiwei.han@novasbe.pt,
hongs@andrew.cmu.edu,
claudia.soares@fct.unl.pt
}

\begin{document}

\maketitle

\begin{abstract}
Explainable Artificial Intelligence (XAI) is essential for the transparency and clinical adoption of Clinical Decision Support Systems (CDSS). However, the real-world effectiveness of existing XAI methods remains limited and is inconsistently evaluated. This study conducts a systematic PRISMA-guided survey of 31 human-centered evaluations (HCE) of XAI applied to CDSS, classifying them by XAI methodology, evaluation design and adoption barrier. Our findings reveal that most existing studies employ post-hoc, model-agnostic approaches such as SHAP and Grad-CAM, typically assessed through small-scale clinician studies. The results show that over 80\% of the studies adopt post-hoc, model-agnostic approaches such as SHAP and Grad-CAM, and that typical clinician sample sizes remain below 25 participants. The findings indicate that explanations generally improve clinician’ trust and diagnostic confidence, but frequently increase cognitive load and exhibit misalignment with domain reasoning processes. To bridge these gaps, we propose a stakeholder-centric evaluation framework that integrates socio-technical principles and human–computer interaction to guide the future development of clinically viable and trustworthy XAI-based CDSS.
\end{abstract}

\section{Introduction}
\label{sec:introduction}

\begin{figure*}[h!]
\centering
\begin{tikzpicture}[scale=1.4, every node/.style={align=center}]

\tikzset{
    set/.style = {
        circle, minimum size=6cm, draw=black, thick, text opacity=1, 
        text width=4cm, align=center
    },
    healthcare/.style = {set, fill=red, fill opacity=0.2},
    xai/.style        = {set, fill=blue, fill opacity=0.2},
    hci/.style        = {set, fill=green, fill opacity=0.2}
}

\node[healthcare,label=above:{\textbf{Healthcare}}] (A) at (0,0) {
    \textbf{CDSS (Sec. \ref{sec:CDSS_general}):}\\
    AI-driven\\
};

\node[xai,label=above right:{\textbf{Explainable AI}}] (B) at (4,0) {
    \textbf{Taxonomy (Sec. \ref{sec:XAI_methods}):}\\
    Type\\
    Dependency\\
    Scope\\
    Output
};

\node[hci,label=below:{\textbf{Human-Centered Evaluations }}] (C) at (2,-3.2) {
    \textbf{Evaluation (Sec. \ref{sec:HCE}):}\\
    Application-grounded\\
};

\begin{scope}
\clip (A) circle (3cm);
\clip (B) circle (3cm);
\fill[yellow!90!orange, opacity=0.9] (C) circle (3cm);
\end{scope}

\node at (2,-0.8) [align=center, text width=3.5cm, font=\bfseries, text opacity=1] {
    Focus of this \\ Review \\ Sec. \ref{sec:papers_list}
};

\end{tikzpicture}
\caption{The survey exclusively focuses on works at the intersection of Healthcare, Explainable AI, and Human-Centered Evaluations in the yellow region}. \label{fig:venn_diagram}
\end{figure*}

Artificial Intelligence (AI) is increasingly being integrated into Clinical Decision Support Systems (CDSS) to improve timely diagnosis, optimize treatment plans, and reduce healthcare costs \cite{sendak2020human}. Estimates suggest that AI adoption in healthcare could reduce U.S. healthcare expenditures by 5–10\%, translating to $200–$360 billion in annual savings while maintaining quality and access \cite{sahni2023potential}. The rapid growth of AI-driven CDSS has also attracted significant investment, with venture capital funding in AI for health exceeding \$11 billion in 2024 \cite{goldsack2024healthcareai}. These trends indicate that the role of AI in medical decision-making will continue to expand \cite{lorenzini2023artificial}.

A key challenge accompanying this expansion is the need for trust and transparency in AI-driven medical decisions \cite{qin2024personalization}. Although deep learning and other black-box models achieve high accuracy, clinicians remain hesitant to rely on AI-generated recommendations \cite{longoni2019resistance,gaczek2023overcoming}. Trust in CDSS is critical because diagnostic errors can have life-threatening consequences \cite{zhang2023ethics}. Explainable AI (XAI) has therefore emerged to enhance transparency, allowing medical professionals to understand model reasoning \cite{loh2022application}. Regulatory frameworks such as the General Data Protection Regulation (GDPR) further reinforce the requirement for explainability in automated decisions.

Despite extensive research in XAI for healthcare \cite{loh2022application}, a critical gap remains: limited systematic evidence exists on how XAI explanations influence real-world clinical adoption and decision-making \cite{ghassemi2021false}. Prior studies have largely emphasized algorithmic innovation over evaluation of clinical usability \cite{amann2022explain}. Moreover, assessments of XAI effectiveness remain fragmented, ranging from proxy simulations to small-scale clinical trials \cite{doshi2017towards}. This raises a central question: \textit{How should XAI techniques be evaluated to ensure usability, trustworthiness, and effectiveness in clinical settings?}


To address this gap, this paper presents a systematic survey of Human-Centered Evaluations (HCE) of XAI methods in Clinical Decision Support Systems (CDSS). Unlike prior surveys on XAI in healthcare \cite{antoniadi2021prediction,loh2022application,Kim2024}, which emphasize algorithmic innovations or cross-domain interpretability frameworks, our focus is on how explanations are assessed with clinicians and other stakeholders in practice. Specifically, we 
(1) categorize existing works based on XAI methodologies, evaluation frameworks, and clinical adoption challenges, offering a structured synthesis of the field;
(2) identify key gaps in evaluation practices, highlighting inconsistencies and challenges such as high cognitive load and misalignment with clinical knowledge;
(3) introduce a conceptual framework to bridge the socio-technical gap in AI-driven CDSS by aligning XAI evaluation methods with diverse stakeholder needs (e.g., clinicians, administrators, and AI developers); and
(4) propose a stakeholder-centric approach for CDSS development, emphasizing the design of CDSS as human-augmentation tools rather than replacements.

%
%
To theoretically ground the misalignment between human expectations and technological capabilities, we draw on Ackerman’s socio-technical gap theory, which highlights the challenges created by such discrepancies \cite{ackerman2000intellectual}. This perspective informs our proposed framework, emphasizing stakeholder-driven evaluations to ensure that CDSS solutions are clinically relevant, interpretable, and effectively integrated into healthcare workflows. By aligning technical advancements in XAI with rigorous human-centered evaluations, we aim to steer research toward developing trustworthy and deployable CDSS solutions.

The remainder of this paper is organized as follows. Section \ref{sec:background} provides background on CDSS, XAI methods, and human-centered evaluations. Section \ref{sec:papers_list} presents a structured review of existing works, categorizing them by XAI methodologies, evaluation frameworks, and clinical domains. Section \ref{sec:challenges} discusses the socio-technical challenges and introduces a conceptual framework to bridge the gap between XAI methodologies and stakeholder needs, concluding with directions for future research.




\section{Clinical Decision Support Systems and Explainable AI Taxonomy} \label{sec:background}

In this section, we provide a brief overview of CDSS, present a taxonomy on XAI, detailing criteria based on type, dependency, and scope; review related work on XAI in healthcare, pointing out a lack of human-centered evaluation; and, in turn, review how such evaluations can be categorized and leveraged. Fig.~\ref{fig:venn_diagram}  shows the focus of this review.

\subsection{Clinical Decision Support Systems (CDSS)} \label{sec:CDSS_general}
CDSS have long been integral to medical decision-making, assisting clinicians in enhancing patient outcomes. Initially emerging in the late 1950s with rule-based systems \cite{ledley1959reasoning}, CDSS evolved significantly with advancements in AI. Modern CDSS now leverage sophisticated machine learning algorithms to process vast datasets efficiently \cite{perez2015bigdata}. Concurrently, the rise of XAI has enabled the integration of interpretable solutions, essential for fostering trust and promoting the adoption of CDSS in clinical practice \cite{ghassemi2021false}.


Operationally, CDSS utilize patient data to generate specific treatment recommendations~\cite{musen2021clinical}. The back-end employs machine learning algorithms trained for specific tasks, processing data from \textit{Electronic Health Records} (EHRs). EHRs include both structured data (\textit{e.g.}, tabular data) and/or unstructured data (\textit{e.g.}, text, images), representing comprehensive digital versions of patients’ medical histories, including diagnoses, treatments, and test results. System performance is evaluated using machine learning metrics pertinent to the specific task. 
In parallel, the front-end runs a user interface (UI) connected to the back-end that clinical practitioners are supposed to interact with. Such UI displays patient data, alerts, recommendations (sometimes with explanations), and decision-support reports in an accessible manner to facilitate clinical decision-making.
To this end, human-centered evaluations are conducted to assess the overall effectiveness and efficiency of the system in the specific task. 
However, researchers have identified that the absence of such evaluations is a primary factor in the lack of adoption of AI-based CDSS solutions \cite{musen2021clinical}, particularly due to insufficient consideration of clinicians’ workflows and the collaborative nature of clinical practice \cite{kohli2018cad,musen2021clinical,wears2005computer,horsky2012interface}.
In this paper, we place emphasis on reviewing studies that actively incorporate human-centered evaluations to ensure alignment with clinical practices.

\subsection{A Taxonomy of XAI Methods} \label{sec:XAI_methods}
Many taxonomies have been proposed to categorize XAI methodologies (\textit{e.g.}, see~\cite{speith2022review}).
In this paper, we present a \textit{conceptual} taxonomy based on the following macro criteria (for a more comprehensive review, please refer to~\cite{molnar2022}):
\begin{itemize}
    \item \underline{\textbf{Type}}: \textbf{Intrinsic} (or Ante Hoc) vs \textbf{Post Hoc};
    \item \underline{\textbf{Dependency}}: \textbf{Model Specific} vs \textbf{Model Agnostic};
    \item \underline{\textbf{Scope}}: \textbf{Local} vs \textbf{Global};
\end{itemize}

\subsubsection{Type: Intrinsic vs Post Hoc}
\textbf{Intrinsic} interpretability is associated with machine learning models such that humans can easily trace their decision-making process. This is typically achieved through common white-box models such as linear models, in which coefficients can be considered the marginal effects on the output in regression problems, and tree methods as decision trees, in which their hierarchical structure allows for direct observation and understanding of the decision-making process. In addition, ensemble models such as random forests or boosting methods, although considered as black box, contain an intrinsic interpretable component in terms of feature contribution. 
Finally, some deep learning models may have intrinsic interpretability components in their structure, \textit{e.g.}, attention weights in Transformers, highlighting the importance of different input features.
\underline{\textit{Methods}}: Linear Models, Tree-models, Attention Weights in Transformers, Bayesian Networks, among others.


\textbf{Post Hoc} interpretability refers to methodologies used after model training, providing insights into how predictions are made without altering the model itself. These methodologies are not intrinsic to the model, requiring further optimization to extract interpretation-oriented representations such as feature importance scores, partial dependence plots, surrogate models, among others.  
Notwithstanding, these methods can be applied on top of intrinsically interpretable models to provide a further layer of explainability. 
\underline{\textit{Methods}}:
 SHAP \cite{lundberg2017shap}, LIME \cite{ribeiro2016should}, GradCAM \cite{selvaraju2017grad}, DeepTaylor \cite{montavon2017explaining}, Integrated Gradients \cite{sundararajan2017axiomatic}, Partial Dependence Plots (PDP) \cite{friedman2001greedy}, Individual Conditional Expectations (ICE) \cite{goldstein2013peeking}, Counterfactual Explanations \cite{mothilal2020dice}, Surrogate Models, among others.


\subsubsection{Dependency: Model Specific vs Model Agnostic}

\textbf{Model Specific} methods are limited to certain families of model architectures, and their use is coupled to the inner workings of those models. For example, neural networks require specific explainability methodologies, \textit{e.g.}, saliency maps such as  GradCAM for CNNs. 
Importantly, note that intrinsic interpretable models are by definition model specific, \textit{e.g.}, interpreting linear regression coefficients is specific to that kind of model only. 
\underline{\textit{Methods}}: Architecture-specific saliency maps (\textit{e.g.}, CAM \cite{zhou2016learning}, GradCAM \cite{selvaraju2017grad}), Integrated Gradients \cite{sundararajan2017axiomatic},  DeepTaylor\cite{montavon2017explaining}, 
among others.

\textbf{Model Agnostic} methodologies are suitable for multiple families of machine-learning models, regardless of their internal structure. These methods are called ``agnostic" because they do not rely on the internal workings of the model, which they treat as black box, but focus on the inputs and outputs only.
These methodologies are applied post-hoc. \textit{\underline{Methods}}: SHAP \cite{lundberg2017shap}, LIME \cite{ribeiro2016should}, PDP \cite{friedman2001greedy}, Accumulated Local Effects (ALE) \cite{apley2020visualizing}, Counterfactual Explanations \cite{mothilal2020dice}, Scoped Rules \cite{ribeiro2018anchors}, among others.


\subsubsection{Scope: Local vs Global}
\textbf{Local} interpretability methods explain an \textit{individual} prediction, \textit{e.g.}, how each feature contributes to a given output, thus approximating model's behavior locally. Also, local methods are particularly suitable for models leveraging unstructured data, as, for instance, they can pinpoint the specific pixels in an image or tokens in a text that predominantly drive predictions. 
\underline{\textit{Methods}}: ICE \cite{goldstein2013peeking}, LIME \cite{ribeiro2016should}, SHAP \cite{lundberg2017shap}, Counterfactual Explanations \cite{mothilal2020dice}, Integrated Gradients \cite{sundararajan2017axiomatic}, Attention Weights, among others. 

\textbf{Global} interpretability methods provide a wider perspective on how a model is explainable across the whole dataset (usually the test set). These methods are particularly suitable for models leveraging structured data. 
While local interpretability is essential for justifying individual choices, global interpretability promotes global model transparency.
\underline{\textit{Methods}}: SHAP \cite{lundberg2017shap}, PDP \cite{friedman2001greedy}, ALE \cite{apley2020visualizing}, Global Surrogates, among others.

\subsection{Related Work on XAI Methods in Healthcare}

We briefly overview existing literature using the XAI methods mentioned in the taxonomy previously discussed.   

SHAP emerged as one of the most widely adopted XAI technique, utilized for identifying critical clinical features in disease prediction and patient outcome assessment \cite{liu2021need,shi2022explainable,chen2021forecasting}. 
The technique proved especially valuable for hospital management applications, including mortality prediction in intensive care units, hospital readmission risk assessment, and surgical complication forecasting, with clinical feature importance analysis revealing consistent patterns where age, laboratory values, and vital signs were frequently identified as primary predictors \cite{nguyen2021prediction,zeng2021explainable,zhang2021explainable}.

Similarly, GradCAM has been widely used for visual explanation tasks, focusing on medical imaging interpretation \cite{figueroa2022interpretable,xu2021clinical,chetoui2021explainable}. 
Also, this technique proved particularly effective for COVID-19 detection from chest X-rays, with studies demonstrating its utility in highlighting lung discriminative regions for pandemic-related diagnosis \cite{shi2021covid,ozturk2020automated}. GradCAM implementations utilized CNN architectures such as ResNet and VGG, extending successfully beyond radiological applications to ophthalmological imaging, dermatological lesion analysis, and histopathological examination \cite{yoo2021xecgnet,thakoor2021robust}.

Other XAI techniques have been applied in more specialized contexts, with LIME adopted for single-instance explanations, Layer-wise Relevance Propagation (LRP) utilized for heatmap generation, and various other methods including fuzzy classifiers employed for specific clinical tasks \cite{binder2021morphological,sabol2020explainable}. Finally, XAI applications in clinical NLP showed significant potential for medical coding automation and mental health assessment
\cite{dong2021explainable,hu2021explainable}. 

However, despite a general positive trend of XAI adoption for clinical usage, those systems fail to consider human feedback as an evaluation principle, leaving socio-technical gaps between clinical practitioners and technological solutions, limiting CDSS effectiveness for an effective patient-centric care. 
Motivated by that, in the next subsection, we provide a review on human-centered XAI evaluations. 


\subsection{Human-Centered XAI Evaluations (HCE)} \label{sec:HCE}

In practical medical applications, socio-technical gaps may arise between the CDSS explainability factors provided by XAI techniques and end-users' perceptions of their utility~\cite{ackerman2000intellectual}. Human-centered evaluations offer methodologies to bridge this gap, aiming to align explanations with user expectations. 
Such evaluations should hopefully reflect human desired properties of explanations, serving some practical end goal~\cite{liao2023rethinking}.
Human-centered evaluations of XAI systems can be categorized into three levels: (1) application-grounded, (2) human-grounded, and (3) proxy evaluation \cite{doshi2017towards}.

\textbf{Application-grounded} evaluations involve testing real tasks with real experts and should be employed whenever possible because of their fidelity to real-world scenarios. For example, in the medical domain, physicians may be evaluated on specific application tasks. Here, domain experts are required for an effective evaluation. For this reason, application-grounded evaluation may be costly to implement.
On the other hand, \textbf{human-grounded} evaluations use proxy tasks with human evaluations, conducting simplified human-subject experiments. This approach is particularly suitable for assessing general aspects of a system. As such evaluations can be performed with layman participants instead of domain experts, they are not advisable in the medical domain.
Finally, \textbf{proxy} evaluations assess proxy tasks without human participation, relying on simulated tasks to provide feedback about a system. 

Similar to human-grounded evaluations, proxy evaluations are not advisable for the full evaluation of CDSS ready for production. However, they may be useful for an early-stage prototype, provided a careful application-grounded evaluation is eventually performed.
These three levels of XAI evaluations imply a trade-off between fidelity and cost, where application-grounded evaluations provide the most realistic insights but are expensive, human-grounded evaluations offer a balance of generalizability and affordability but lack domain specificity, and proxy evaluations are the most scalable and cost-efficient but least reflective of real-world performance.
In the medical domain, application-grounded evaluation is the preferred choice because it ensures high-fidelity testing with domain experts. This is crucial for patient safety and clinical reliability, despite its higher cost and complexity.

Human-centered evaluation methodologies leverage various techniques, either qualitative or quantitative. In a nutshell, the most widely known are: (1) Think-Aloud (TA) studies, (2)  Interviews (I), and (3) Surveys (S). 
First, in think-aloud studies participants verbalize their thoughts while performing tasks (concurrent) or afterward (retrospective), which can be very useful in qualitatively explaining cognitive processes and identifying usability issues. 
Second, interviews are a qualitative way to gather detailed information from users about their experiences with a system. They are versatile tools in all stages of design, ranging from open-ended unstructured interviews to tightly controlled structured interviews, with semi-structured interviews offering a valuable compromise. A variant of those is Focus Groups (FG), in which a small number of participants are gathered to evaluate their opinions about a product or service. 
Finally, survey methodologies mainly rely on quantitative questionnaires that help researchers gather data on users’ satisfaction, usability issues, and interaction patterns.

\section{Dissecting existing HCE-XAI in CDSS} \label{sec:papers_list}


\begin{figure*}[h!]
    \centering
    \includegraphics[width=0.7\textwidth]{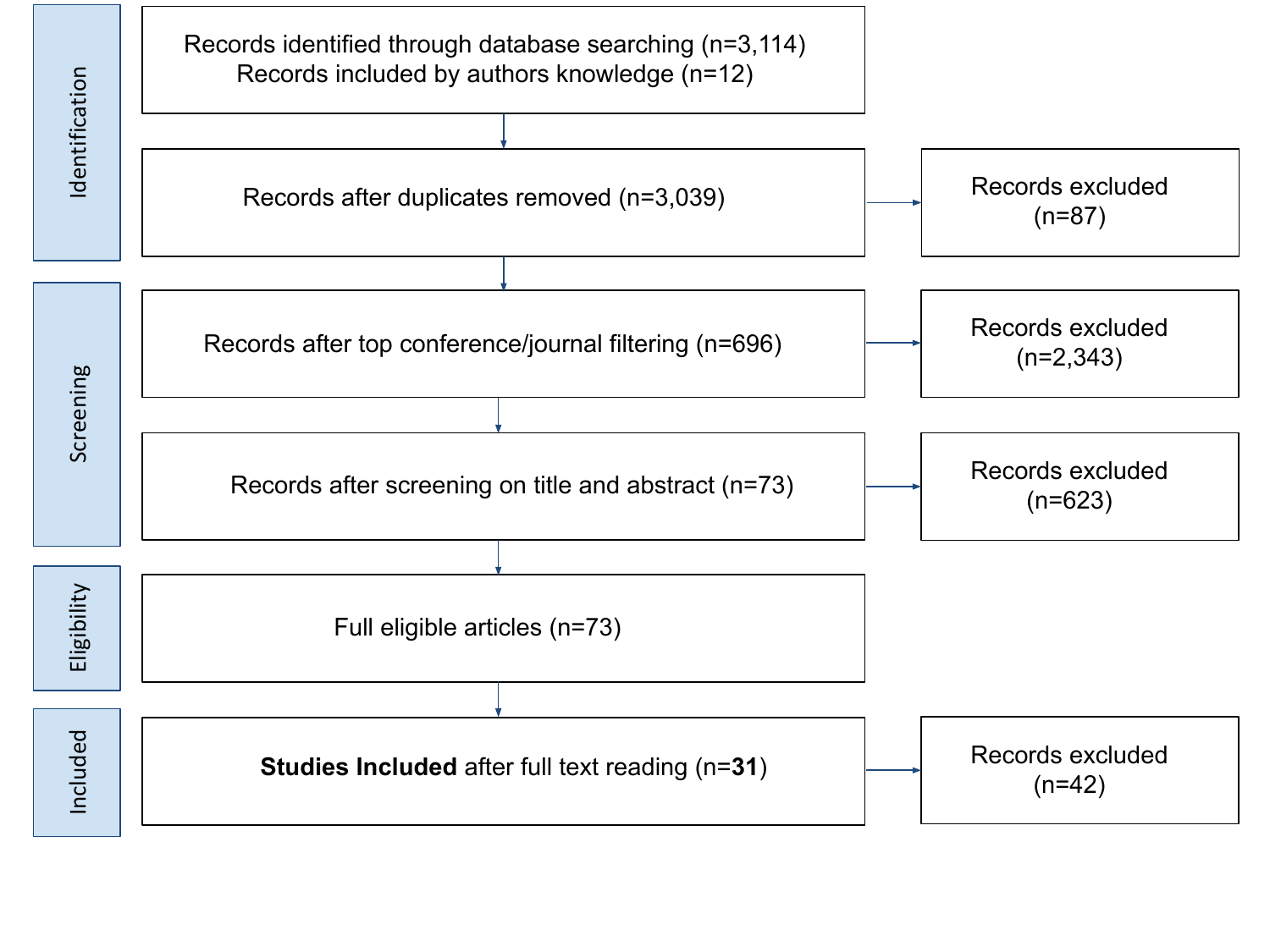}
    \caption{PRISMA Flow Diagram detailing the systematic process of identifying, screening, and selecting studies for inclusion.}
    \label{fig:PRISMA}
\end{figure*}

\begin{figure}[h!]
    \centering
    \includegraphics[width=0.45\textwidth]{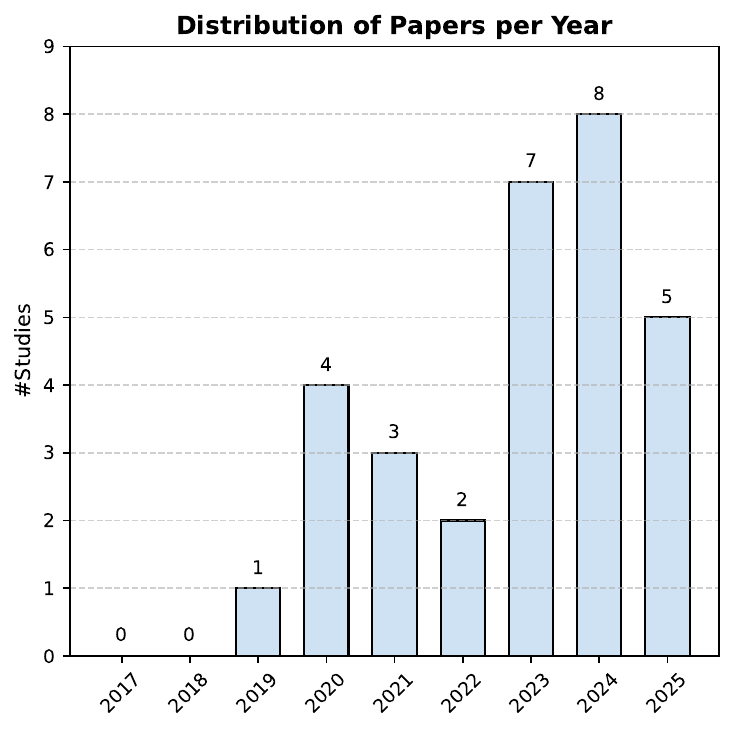}
    \caption{Histogram of retrieved papers per year. 
    }
    \label{fig:years}
\end{figure}

We adopt the PRISMA methodology to form a database of possible candidate research papers. 
Fig.~\ref{fig:PRISMA} shows the PRISMA flow diagram. 
We query three databases to cover research in computer science, human-computer interaction and healthcare: ACM Digital Library, Web of Science, and PubMed. 
Our strategic keyword selection is as follows. We start with simple keywords such as \textit{Clinical Decision Support System (CDSS)}, \textit{Explainable AI (XAI)}, \textit{Healthcare}, and \textit{Human-Centered Evaluations (HCE)}, among others, and we iteratively refine the selection by adding and deleting more granular terms to capture a broader scope (Appendix \ref{sec:appendix_keywords}).
We then narrow the selection of research papers down considering the following criteria:
(1) top-tier conference venue (CORE A* or A) or SJR Q1 journal (\textit{PRISMA ``Screening"});
(2) publication date after 2017, motivated by the DARPA XAI Program announced in May 2017 \cite{gunning2019darpa}, which fostered general innovation in XAI (\textit{PRISMA ``Screening"}), 
(3) explicit clarity to understand which XAI and HCE methodologies were adopted, as well as the medical field. Here, each XAI/HCE methodology would at least fit into either one of the XAI taxonomy or HCE methods, respectively, as outlined in the previous section (\textit{PRISMA ``Included"}). 
As a result, we identified 31 papers encompassing the three macro-categories as shown in Fig.~\ref{fig:venn_diagram}.
In Table \ref{tab:tab2}, we provide a summary of the papers retrieved. While, in Fig.~\ref{fig:years}, we preliminary show the distribution of retrieved papers per year of publication. Here, with a general positive trend, most of the papers were authored in 2023 and 2024.

We structure the discussion of the papers as follows. 
Firstly, we start examining CDSS adopting intrinsic interpretable machine learning models (Section \ref{sec:cdss_intrinsic_only}). As the explainability evaluation of these models is endogenous, it is relatively straightforward to trace back how a prediction could be made. 
On the contrary, post hoc methodologies require further optimization to extract interpretable information. Therefore, we separately examine CDSS using post hoc explainability methodologies (Section \ref{sec:cdss_post_hoc_intrinsic}). 
Finally, we discuss CDSS adopting XAI methodologies in deep learning (Section \ref{sec:cdss_deep_learning}). Deep learning models usually integrate unstructured data (text, images), meaning that different architectures are required, translating also into different explainability needs. For example, we will document that most of these models rely on computer vision model architectures, often requiring specific and different explainability methodologies with respect to the ones discussed in Section \ref{sec:cdss_post_hoc_intrinsic}.
As a last remark, we group papers by medical field, to facilitate discussion and scrutinize whether some explainability and evaluation patterns emerge. Ethical approval was not required for this study, as it is a systematic review of previously published literature and does not involve the collection of new human or animal data.

\begin{table*}
    \centering
    \small
    \caption{Summary of Papers by ML Models, XAI Methods, XAI Taxonomy, HCE Methodologies, Medical Fields, and Clinician Perceptions.}\label{tab:tab2}
    \begin{tabular}{|l|l|l|l|l|l|l|}
        \hline
         \textbf{Paper} & \textbf{Field} & \textbf{Model(s)} & \textbf{XAI(s)} & \textbf{XAI Taxonomy Sec. \ref{sec:XAI_methods}} & \textbf{HCE(s)} & \textbf{Perc.} \\
        \hline      
        
        \cite{barda2020qualitative} & Critical Care & RF & SHAP & PH, AGN, LOC & FG & $\uparrow$ \\

        \cite{brennan2019comparing} & Critical Care & GAM, RF, + & - & INT, LOC & TA & $\uparrow$ \\


        \cite{ellenrieder2023promoting} & Radiology & CNN & I. Gradients & PH, SPEC, LOC & TA & $\uparrow$ \\

        \cite{hwang2022clinical} & Sleep Medicine & CNN & Saliency & PH, INT, LOC & TA, I & $\rightarrow$ \\


        \cite{kumarakulasinghe2020evaluating} & Critical Care  & RF & LIME & PH, AGN, LOC & S & $\uparrow$ \\

        \cite{kovalchuk2022three} & Endocrinology & RF, XGB, + & SHAP & PH, AGN, LOC & S, I, + & $\rightarrow$ \\ 


        \cite{matthiesen2021clinician} & Cardiology & RF & LIME & PH, AGN, LOC & I & $\rightarrow$ \\

        \cite{neves2021interpretable} & Cardiology & KNN, CNN, + & LIME, + & PH, AGN, LOC & S & $\rightarrow$  \\

        \cite{pumplun2023bringing} & Radiology & CNN & I. Gradients & PH, SPEC, LOC & S & $\uparrow$ \\ 

        \cite{rajashekar2024human} & Gastroenterology & RF & PDP, ICE, ALE & PH, AGN, LOC/GLOB & I, S, + & $\uparrow$ \\

        \cite{sabol2020explainable} & Oncology & CNN & X-CFCMC & PH, AGN, LOC & S & $\uparrow$ \\ 

        \cite{singh2021evaluation} & Ophthalmology & CNN & DeepTaylor & PH, AGN, LOC & S & $\uparrow$ \\

        \cite{sivaraman2023ignore} & Critical Care  & XGB & SHAP & PH, AGN, LOC  & TA & $\rightarrow$ \\

        \cite{zhang2024rethinking} & Critical Care & LSTM & Attention & INT, LOC & I & $\rightarrow$ \\ 

        \cite{Bienefeld2023} & Critical Care & Tree  & SHAP & PH, AGN, LOC & S, FG+ & $\rightarrow$ \\


        \cite{bhattacharya2023directive} & Endocrinology & Logit & SHAP & PH, AGN, LOC & I, S+ & $\rightarrow$ \\ 

        \cite{abraham2023integrating} & Critical Care & - & SHAP & PH, AGN, LOC & I & $\uparrow$ \\


        \cite{cabitza2025fromoracular} & Radiology & CNN & CAM & PH, SPEC, LOC  & S & $\rightarrow$ \\ 

        \cite{chanda2024dermatologist} & Dermatology & CNN & GradCAM & PH, SPEC, Local  & S  & $\uparrow$ \\

        \cite{chari2023informing} & Endocrinology & BERT & SHAP & PH, AGN, LOC & TA & $\uparrow$ \\ 


        \cite{famiglini2024evidence} & Radiology & CNN & CAM & PH, SPEC, LOC & S & $\uparrow$ \\ 

        \cite{gombolay2024effects} & Neurology & DT & - & INT, LOC & S & $\rightarrow$ \\

        \cite{gu2020acase} & Oncology & XGBoost & SHAP & PH, AGN, LOC & S & $\uparrow$ \\

        \cite{he2024vms} & Critical Care & DT, RF, + & SHAP & PH, AGN, LOC/GLOB & TA & $\uparrow$ \\

        \cite{hur2025comparison} & Critical Care & Boosting & SHAP & PH, AGN, LOC & S & $\uparrow$ \\

        \cite{ihongbe2024evaluating} & Radiology & CNN & LIME, + & PH, AGN/SPEC, LOC & S & $\uparrow$ \\

        \cite{jing2025development} & Geriatrics & XGB & SHAP & PH, AGN, LOC & S & $\uparrow$ \\

        \cite{jung2025evaluating} & Critical Care & DT, RF, + & SHAP & PH, AGN, LOC & I & $\uparrow$ \\

        \cite{kayadibi2025ai} & Dentistry & CNN & LIME & PH, AGN, LOC & S & $\uparrow$ \\

        \cite{rainey2024operationalizing} & Radiology & CNN & GradCAM & PH, SPEC, LOC & S & $\rightarrow$ \\

        \cite{singla2023medical} & Radiology  & CNN & Counterf. & PH, SPEC, LOC & S & $\uparrow$ \\

        \hline
    \end{tabular}
\vspace{2mm}
\begin{minipage}{0.9\textwidth}
\footnotesize
\textit{Note}—``$\uparrow$" indicates a Positive clinician perception; ``$\rightarrow$" indicates a Neutral perception; ``$\downarrow$" indicates a Negative perception. \\
In Model, XAI and HCE, ``+" indicates more methods used. 
Abbreviations: FG: Focus Group; TA: Think-Aloud; I: Interview; S: Survey; \\ 
RF: Random Forest; CNN: Convolutional Neural Network; XGB: XGBoost; DT Decision Tree; Logit: Logistic Regression; GAM: Generalized Additive Model; KNN: K-Nearest Neighbor; LSTM: Long Short-Term Memory; BERT: Bidirectional Encoder Representations from Transformers; \\ 
PH: Post Hoc; AGN: Agnostic, SPEC: Specific; INT: Intrinsic; LOC: Local; GLOB: Global.
\end{minipage}
\end{table*}

\subsection{CDSS adopting Intrinsic Interpretable Models} \label{sec:cdss_intrinsic_only}

Intrinsic interpretable models are easily testable in the medical field for CDSS due to their transparency, as no post-hoc evaluations are necessarily required to perform model evaluations. Examples in this scope are the works from  Brennan et al.~\cite{brennan2019comparing} in \textbf{Critical Care} medicine,  
and Gombolay et al.~\cite{gombolay2024effects} in \textbf{Neurology}. 
~\cite{brennan2019comparing} validated the usability of \textit{MyRiskSurgery}, an analytical CDSS based on generalized linear models 
and random forests
to estimate postoperative risk complications \cite{bihorac2019mysurgeryrisk}. Through think-aloud studies, a set of physicians evaluated risk complications for a set of 8-10 individual cases in two case studies such as diagnosing acute kidney injury and Intensive Care Unit (ICU) admission, with the system displaying the top three contributing features for each prediction. As a result, physicians experienced notable improvements in their risk assessment, which led to a substantial net improvement in reclassification rates from a no XAI evaluation (+12\% and +16\%, respectively), finding the system to be beneficial. 

Finally,~\cite{gombolay2024effects} conducted a study to evaluate the impact of various XAI techniques on clinician interactions with CSDSS in pediatric and adult neurological disorder. Clinicians were exposed to different XAI methodologies, including decision trees, crowd-sourced agreement, case-based reasoning, probability scores, counterfactual reasoning, feature importance, templated language, and no explanations. 
The study found that decision trees were perceived as the most explainable. Interestingly, increasing neurology experience and perceived explainability were associated with degraded performance, raising the concern that there is not one-size-fits-all XAI medical approach, but strong CDSS personalization is warranted.

\subsubsection*{Short Summary of Findings}
With only two studies surveyed, \textit{MyRiskSurgery} showed positive perception and seamless usability by clinicians \cite{brennan2019comparing}. While, ~\cite{gombolay2024effects} raised the concern that XAI customizations are required for better understandability based on the specific applied case.

\subsection{CDSS adopting Post Hoc XAI Methodologies } \label{sec:cdss_post_hoc_intrinsic} 
Post-hoc methodologies can be applied on top of intrinsic interpretable models to provide a further layer of explainability. 
All these works employed model-agnostic methodologies, mostly SHAP and LIME to explain predictions locally, \textit{i.e.}, at the patient-level.

In the field of \textbf{Critical Care} medicine,
~\cite{barda2020qualitative} developed a CDSS to prevent in-hospital mortality in Pediatric Intensive Care Units (PICU). The CDSS leveraged a random forest and SHAP values on top. 
In general, SHAP methods quantify each feature’s contribution to a model’s prediction by computing Shapley values from cooperative game theory (for an exhaustive review please refer to \cite{lundberg2017shap,shapley1953value}).
Overall, feedback from focus group sessions with clinicians indicated that the proposed solution was perceived as useful. However, the capacity to digest the information varied across clinical roles and their level of predictive modeling knowledge, with preferred solutions that required less cognitive effort.
Next,
~\cite{he2024vms} developed the Visualization for Model Sensemaking and Selection (VMS), an interactive tool designed to aid in the selection of machine learning models to predict patients' Length of Stay (LOS) in ICU. The system employs a combination of decision tree, random forest, LSTM, and GRU models, with SHAP values used to provide both local and global explanations of feature importance. 
Feedback from a user study  study involving 16 medical professionals indicated that the tool enhanced CDSS interpretability, although it was noted that there was a learning curve associated with its use.

With the same goal, and adopting a similar SHAP methodology, ~\cite{jung2025evaluating} 
found that clinicians aligned more closely with the CDSS after receiving explanations, though full alignment was not reached.
Next, ~\cite{hur2025comparison} developed the \textit{Precision Maximum Surgical Blood Ordering Schedule-Thoracic Surgery} (pMSBOS-TS) to predict personalized red blood cell demand in thoracic surgery patients. pMSBOS-TS employs a backbone using boosting methods, integrating SHAP to highlight the top three contributing features, and was evaluated with 63 clinicians, including surgeons and physicians, who participated in a repeated-measures experiment using clinical vignettes to evaluate three explanation formats: Results Only (RO), Results with SHAP visualizations (RS), and Results with SHAP visualizations plus clinical explanations (RSC). The RSC format significantly improved trust, satisfaction, and acceptance compared to the other formats.

Works discussed so far have primarily focused on perioperative or intraoperative critical care. 
As for postoperative care, ~\cite{abraham2023integrating} developed a ML-augmented tool designed to predict complications such as acute kidney injury, delirium, pneumonia, among others. 
The tool employs SHAP values to provide interpretations of risk factors, and was evaluated with 17 clinicians, including anesthesiologists, surgeons, and critical care physicians, who participated in cognitive walkthroughs and interviews to evaluate the tool's usability and integration into clinical workflows. The findings indicated a high level of agreement between ML-generated risk predictions and clinicians' manual risk assessments. Also, clinicians appreciated the tool's ability to streamline report preparation and enhance care planning by highlighting patient-specific risks. 

As for sepsis treatment, ~\cite{sivaraman2023ignore} examined clinicians' interactions with an AI-based interpretable sepsis treatment recommender. The system was modeled using XGBoost, and SHAP as feature explainer. 
Through a think-aloud study involving 24 intensive care clinicians, the analysis revealed that explanations improved clinicians' perceptions of the AI's usefulness. In particular, the study identified four distinct behavior patterns among clinicians summarised as: \textit{Ignore}, \textit{Negotiate}, \textit{Consider}, and \textit{Rely}. Specifically, in the Negotiate group, clinicians displayed a tendency to weigh and prioritize aspects of the AI recommendations, suggesting that treatment decisions in the ICU may involve partial reliance on AI recommendations, with potential implications for the effectiveness of chosen treatments. 
Overall, clinicians who found the AI recommendations most useful saw it as additional evidence alongside their assessments.
Next, in neurocritical care,~\cite{Bienefeld2023} presented the development and evaluation of the \textit{Delayed Cerebral Ischemia Predictor} (DCIP), a CDSS designed to forecast delayed cerebral ischemia in patients with aneurysmal subarachnoid hemorrhage. The system combined Extremely Randomized Trees and SHAP values to display static and dynamic risk contributors, and was evaluated with 112 clinicians and developers using a mixture of surveys, focus groups, and think-aloud studies. Results showed that 
while developers emphasized model interpretability, 
clinicians sought clinical plausibility, leading to design tensions around the utility of SHAP explanations. 

All the works in critical care reviewed so far adopted SHAP as explainability framework. On the contrary,~\cite{kumarakulasinghe2020evaluating} adopted
LIME for sepsis prediction. 
In general, LIME methods approximate a model’s behavior around a specific instance by fitting an interpretable surrogate model to locally perturbed samples \cite{ribeiro2016should}.
Developers trained several models and applied LIME to the best-performing random forest. Physicians evaluated agreement with sepsis predictions and independently ranked key features before revealing LIME explanations. Afterward, they rated satisfaction on a five-star Likert scale. Physicians agreed with the model in 87\% of cases, with LIME aligning on at least two features for 69\%. Overall, 78\% were satisfied and 68\% showed positive attitudes toward explainability, though inconsistencies in LIME limited trust.

Next, in the field of \textbf{Endocrinology},~\cite{kovalchuk2022three} proposed a three-staged CDSS for diabetic diagnosis as follows: (1) \textit{Basic Reasoning}, \textit{i.e.}, implementing healthcare practices by referencing existing policies, (2) \textit{Data-Driven}, \textit{i.e.}, applying machine learning models in the context of the previous stage, and (3) \textit{Explanation}, in which SHAP values were applied on top of (2) to provide better recommendations. 
Unexpectedly, 
the basic reasoning case (1) showed higher understandability, followed by the explanation case (2), and the data-driven case (3). These findings suggest that adding explanations is useful, but well-grounded, known conventional clinical practices may still outperform AI-driven solutions.
Similarly, ~\cite{bhattacharya2023directive} developed a visualization-based CDSS to monitor and evaluate the risk of diabetes onset. The system was built upon a logit model and incorporated multiple explanation types, including rule-based data-centric explanations, SHAP, and counterfactual explanations generated using DiCE \cite{mothilal2020dice}. A preliminary focus group study revealed that while SHAP-based explanations were informative, practitioners found them too complex and instead preferred simpler, data-centric visualizations such as bar charts combined with comparative patient data and actionable suggestions. 
Similarly, counterfactual explanations were valued for their ability to present example-based recommendations, supporting actionable insights. 
In contrast, feature-importance explanations (SHAP) were seen as less actionable. 
Finally, ~\cite{chari2023informing} developed a CDSS to enhance the interpretability of risk predictions for chronic kidney disease (CKD) among type-2 diabetes mellitus (T2DM) patients. The system employs LLMs, specifically BERT and SciBERT, to generate contextual information that align AI predictions with clinical contexts, and integrates SHAP values for deeper understanding of the risk factors involved. The evaluation involved an expert panel of 4 medical professionals who assessed the system's explanations for 20 prototypical patients, highlighting the need to provide actionable insights within clinical workflows.

Across diverse medical domains, we documented a similar trend of how post hoc XAI methods were used. 
%
%
%
For instance, in \textbf{Oncology},~\cite{gu2020acase} proposed a CDSS for breast cancer recurrence prediction that combined XGBoost with SHAP-based visualizations within a case-based reasoning framework. A survey of 32 oncologists revealed a favorable reception, highlighting SHAP’s usefulness compared to alternative interpretability methods. A similar emphasis on SHAP was seen from ~\cite{jing2025development} that, in the field of \textbf{Geriatrics}, developed an interpretable XGBoost model for frailty risk prediction using real-time health data and validated scales. Through a mixed-methods evaluation involving healthcare professionals and data scientists, the study reported strong interpretability (mean rating 4.30/5) and high user satisfaction. In contrast, ~\cite{matthiesen2021clinician}, in \textbf{Cardiology}, developed a CDSS employing a random forest model visualized via LIME to predict ventricular fibrillation risk, finding that while the explanations increased clinicians’ confidence, they did not alter clinical decisions. 
Finally, in \textbf{Gastroenterology},~\cite{rajashekar2024human} integrated LLMs with traditional ML by developing  \textit{GutGPT}, combining natural language reasoning with ICE and ALE plots to generate interpretable narratives for gastrointestinal bleeding risk prediction. 
Broadly speaking, ICE plots visualize how a model prediction for individual instances changes as a single feature varies.
More globally, ALE plots summarize the average effect of a feature on model predictions by integrating local changes.
Users perceived such explanations positively, suggesting the promise of hybrid approaches that blend human-understandable text with quantitative interpretability.


\subsubsection*{Short Summary of Findings}

Across the 15 studies reviewed, eight in Critical Care, three in Endocrinology, and the remaining in Oncology, Cardiology, Geriatrics, and Gastroenterology, perceptions of XAI-based CDSS were predominantly positive. Twelve studies employed SHAP, and two used LIME for patient-specific assessments of feature contributions. This functionality, particularly in time-sensitive domains such as critical care, appeared valuable in augmenting clinicians’ confidence and perceived reliability of AI-driven recommendations.
Among the studies reporting positive perceptions, explanations were generally valued for their clarity, ease of interpretation, and perceived cognitive support in clinical reasoning \cite{barda2020qualitative,kumarakulasinghe2020evaluating,gu2020acase,jing2025development,rajashekar2024human}. 

Several works emphasized that despite the favorable attitude, certain limitations persisted, including the need for user adaptation, indicating a learning curve associated with use \cite{he2024vms}, as well as for solutions demanding less cognitive effort \cite{barda2020qualitative}.
Finally, neutral perceptions typically reflected a cautious attitude toward integration, where explanations were appreciated as complementary evidence rather than decisive inputs \cite{sivaraman2023ignore,matthiesen2021clinician,kovalchuk2022three}. While interpretability was recognized as beneficial, traditional heuristics and clinical experience often retained precedence. Moreover, studies reported that explanation formats varied in their perceived actionability, in which counterfactual and data-driven ones being seen as more actionable than feature-importance visualizations such as SHAP \cite{bhattacharya2023directive}.
In aggregate, the evidence suggests a consistently positive or neutral reception of XAI in clinical decision support, with no reported negative perceptions.

\subsection{CDSS adopting XAI Methodologies in Deep Learning} \label{sec:cdss_deep_learning}

CDSS using deep learning mainly employ computer vision architectures to generate localized explanations of their predictions, most commonly represented as visual saliency maps or heatmaps highlighting the regions of interest that influenced the model’s output.
As we will see, most of these CDSS are applied in \textbf{Radiology}, as it is the field where 
medical imaging data, in the form of X-ray, Computed Tomography (CT) as well as non-ionizing methods like Magnetic Resonance Imaging (MRI), could be leveraged by deep learning models for diagnostic inference. 

In these fields,~\cite{ellenrieder2023promoting} extended the work of~\cite{pumplun2023bringing} that leveraged a CNN and the integrated gradients framework to produce heatmaps of image regions to highlight cancer malignancy probabilities. 
In technical terms, the integrated gradients method attributes a model’s prediction to its input features by accumulating the gradients of the output with respect to the input 
from a baseline to the actual input, producing a fine-grained, axiomatic feature attribution map.
%
In collaboration with radiologists, a think-aloud study evaluated learning outcomes across four UI designs: non-explainable high-performing (NEHP), explainable high-performing (EHP), non-explainable low-performing (NELP), and explainable low-performing (ELP). The results confirmed four hypotheses: (H1) NEHP yielded modest learning gains; (H2) EHP produced greater gains; (H3) NELP induced false learning; and (H4) ELP mitigated false learning, showing that explainability may significantly learning. 
%
%

Next, using CAM methods, 
~\cite{cabitza2025fromoracular} proposed a Judicial AI, a framework
focusing on diagnosing vertebral fractures using X-ray images, employing CAM activation maps as primary explainability technique. 
In general, CAM methods identify the spatial regions within an input image that most contribute to a model’s prediction by linearly combining convolutional feature maps weighted by class-specific coefficients, typically obtained from architectures employing global average pooling. These maps highlight anatomical regions supporting different diagnostic option. The system was evaluated through a survey involving 16 medical professionals, including spine surgeons and radiologists, who assessed 18 complex X-ray cases. Results indicated that Judicial AI improved diagnostic accuracy and clinician confidence, particularly among experienced users, while less experienced users found the system more cognitively demanding.
Similarly,~\cite{famiglini2024evidence} conducted a study to evaluate the impact of CAM in support to a CDSS using a CNN to detect thoracolumbar fractures from X-rays. 
Their evaluation through an online questionnaire administered to 16 medical professionals, including spine surgeons and musculoskeletal radiologists, 
indicated that traditional coloring schemes in CAM consistently yielded higher diagnostic accuracy and increased physician confidence. 

Next, GradCAM methods have also been used for similar tasks. 
Broadly speaking, Grad-CAM methods generalize CAM by using the gradients of the target output with respect to feature maps to derive these weights, allowing class-discriminative localization in convolutional architectures without requiring architectural modifications.
In this subfield, 
~\cite{rainey2024operationalizing} conducted a study to evaluate the impact of a CDSS 
to analyze musculoskeletal radiographs. The AI behind provided two types of feedback: binary diagnoses and visual explanations using GradCAM. The study involved 12 reporting radiographers, who were presented with 18 complete examinations. The evaluation focused on the participants' trust in the AI system and their propensity to change decisions based on AI feedback, indicating a significant correlation between agreement with AI feedback and trust, particularly with the GradCAM visualizations. However, participants showed a preference for simple binary feedback over heatmaps, as the latter often led to disagreement.

Always in radiology, 
~\cite{singla2023medical} proposed a Generative Adversarial Network (GAN)-based CDSS for X-ray images dissection with multiple explanation frameworks.
Broadly speaking, GANs are a class of machine learning models in which two neural networks, the generator and the discriminator, are trained simultaneously in a competitive framework to produce increasingly realistic synthetic data. 
The CDSS was evaluated through a human-centered study involving 12 diagnostic radiology residents, who assessed the understandability, decision justification, visual quality, identity preservation, and overall helpfulness of the explanations. The results indicated that the counterfactual explanations significantly improved the participants' understanding of the classifier's decisions compared to other methods like saliency maps, suggesting that visual explanatory formats are not always better off than more standard approaches. 
Next, with a more comprehensive evaluation of XAI techniques, 
~\cite{ihongbe2024evaluating} developed a CDSS based on two CNNs to diagnose pneumonia from chest X-rays and CT scans, employing 
GradCAM and LIME to provide explanations. A human-centered evaluation was conducted through surveys involving medical professionals, assessing the XAI tools in terms of clinical relevance, revealing a generally positive perception of XAI systems among participants, although there was a noted lack of awareness regarding their practical applications. Notably, GradCAM was preferred over LIME for its superior performance. 


Next, in \textbf{Oncological Imaging},~\cite{sabol2020explainable} adopted a \textit{Cumulative Fuzzy Class Membership Criterion} (CFCMC) to predict colorectal cancer. The CDSS was based on two segmentation systems for whole-slide histopathological images: one using a standalone CNN and the other using the CFCMC classifier on CNN-extracted features to provide explanations.
Through surveys with 14 pathologists, the CFMC showed higher acceptability compared to a plain CNN, pointing out that explainable classifiers improve AI usability and reliability in medical settings.
Finally, in 
\textbf{Dental Radiology},
~\cite{kayadibi2025ai} proposed an innovative \textit{Explainable Mandibular Third Molar Convolutional Neural Network} (E-mTMCNN) architecture to enhance the detection of mandibular third molars in panoramic radiography, aiming to improve treatment in dentistry. E-mTMCNN utilized a CNN model, equipped with LIME to provide localized, interpretable insights into the model's predictions. 
Ten specialized surveyed dentists assessed the system, indicating high clinical acceptance and confidence in the system's reliability.

Finally, across various clinical specialties, studies consistently highlight the importance of human-centered evaluation in determining the utility of XAI methods. In \textbf{Ophthalmology},~\cite{singh2021evaluation} assessed multiple visual explanation techniques for a CNN-based model trained to diagnose retinal diseases using a counterfactual explanatory method, showing positive reception by radiologists. While in \textbf{Cardiology},~\cite{neves2021interpretable} similarly applied various explainability methods to a CDSS based on ECG time series data for predicting heart arrhythmias. In both domains, surveys identified attribution methods, specifically DeepTaylor and LIME, as particularly useful, especially for less experienced clinicians. In \textbf{Critical Care},~\cite{zhang2024rethinking} developed \textit{SepsisLab}, leveraging an attention module (explanatory part) feeding into an LSTM to highlight important variables for sepsis prediction. Here, semi-structured interviews revealed that while explanations were valued, shifting the AI focus away from final decision 
could further strengthen the user experience. 
Next, in \textbf{Dermatology},~\cite{chanda2024dermatologist} implemented GradCAM visualizations for CDSS assisting to diagnose skin melanoma. The system was evaluated with 116 clinicians across multiple phases. Although XAI did not significantly improve diagnostic accuracy, it notably increased clinicians’ confidence and trust. Finally, in \textbf{Sleep Medicine},~\cite{hwang2022clinical} developed a CNN-based CDSS to measure sleep staging, providing explanations through convolutional filters, saliency maps, and intermediate activations. User studies indicated that explanations effectively confirmed AI predictions but were less helpful in determining ambiguous cases, reflecting the clinicians’ desire to validate predictions while relying on their domain expertise. 

\subsubsection*{Short Summary of Findings}
Of the 31 studies, we identified 14 studies leveraging deep learning for CDSS development. CAM-based methods (including GradCAM) were the most adopted XAI framework with 5 occurrences. However, we documented a scattered application of several frameworks ranging from integrated gradients to LIME, with no clear framework to dominate as compared to SHAP for CDSS evaluated in the previous section. 
As for clinicians' perception, the majority proved to helpful with 11 studies showing positive reception \cite{ellenrieder2023promoting,pumplun2023bringing,sabol2020explainable,kayadibi2025ai,singh2021evaluation,neves2021interpretable,chanda2024dermatologist,singla2023medical,ihongbe2024evaluating}. The remaining papers documented neutral XAI perceptions with: ~\cite{cabitza2025fromoracular} finding that less experienced practitioners found explanations cognitively demanding,~\cite{zhang2024rethinking} shifting XAI away from final decision was perceived more reliable, and ~\cite{hwang2022clinical} finding that AI predictions were less helpful in revealing ambiguous cases, reflecting the clinicians’ preference to validate predictions while relying on their domain expertise. 

\subsection*{Aggregate Summary of Reviewed Studies}

\begin{figure*}[h!]
    \centering
    \includegraphics[width=\textwidth]{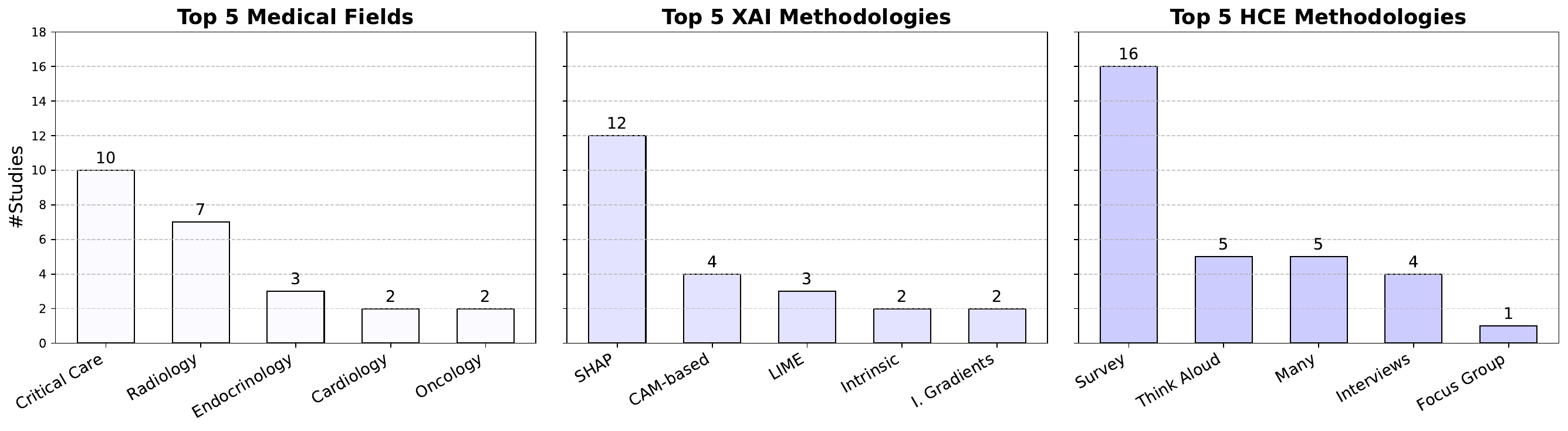}
    \caption{Distribution of top XAI and HCE methodologies and top medical fields where CDSS were used.}

    \label{fig:summary_stats}
\end{figure*}


Fig.~\ref{fig:summary_stats} shows the distribution of XAI and HCE methodologies, and the distribution of the medical fields. We here briefly describe those aggregate findings. 
Across the 31 studies reviewed, the most frequently employed XAI method was SHAP, appearing in 12 papers, followed by CAM-based methods in 4 studies, and LIME in 3 studies. Other methods such as Integrated Gradients, Attention, ICE/ALE, and X-CFCMC were each applied in one or two cases, while a minority either used multiple XAI techniques or intrinsic interpretable models. As expected, all the works use at least one local interpretability method. 

Regarding human-centered evaluation (HCE), surveys represented the predominant method (16 studies), followed by think aloud protocols (5 studies) and interviews (4 studies). Only one study employed the focus group, while 5 studies used mixed/multiple HCE strategies. 
Participant numbers ranged widely, from 3 to 116 subjects, with a median sample size of approximately 16 participants, indicating heterogeneity in evaluation scale. Notably, only 9 studies employed at least 30 participants, questioning the statistical significance validity of the HCE evaluations. This is an inherent limitation. 

In terms of perception outcomes, 20 studies reported a predominantly positive attitude toward the XAI-integrated CDSS, whereas 11 studies were overall neutral. No study reported a negative perception, \textit{i.e.}, XAI was never completely disregarded and seen as a barrier precluding reasonable medical practice. 
Finally, in terms of medical application, Critical Care emerged as the most represented field (10 studies), followed by Radiology (7 studies), with the remaining studies distributed across specialties such as Endocrinology, Cardiology, Oncology, Neurology, and others. 
In this regard, it may be reasonable to expect a high concentration of CDSS studies in Critical Care because this domain generates large volumes of complex, high-frequency data and require rapid, high-stakes decision-making that may benefit from data-driven XAI systems.
With a similar logic, in Radiology, the visual format of the input data is particularly suitable for developing XAI systems, giving the possibility to highlight regions of interests that support model predictions.

\section{Discussion} \label{sec:challenges}

\subsection{Thematic Analysis of XAI-based CDSS} \label{sec:thematic_analysis}

We have reviewed the diversity of CDSS that adopt XAI and human-centered evaluations in the medical field. Given the high-risk environment, application-grounded evaluations with domain experts were the only evaluation methodology. 
On average, clinical perceptions of XAI-based CDSS were mostly positive (20 studies), followed by neutral perceptions (11 studies), reflecting conditional support for XAI, or in other words, recognizing the promise of XAI while expressing doubts about its trust. Notably, no study showed negative perceptions, \textit{e.g.}, XAI viewed as a clear barrier for clinical decision-making. 

Next, we briefly outline the main evaluation themes that emerged, reporting a summary in Table \ref{tab:themes_analysis}.
Firstly, positive perceptions were mostly linked to \textbf{Theme 1: Improvements in Risk Assessments}. Here, most of the works surveyed fall in this theme, \textit{e.g.}, \cite{brennan2019comparing,he2024vms,hur2025comparison,kumarakulasinghe2020evaluating,jing2025development,famiglini2024evidence,kayadibi2025ai}, among others, where XAI was seen as an augmentative assisting tool to clinicians to improve their diagnoses.
In this theme, we identified sub-themes such as: (1.1) \textit{Support for Less Experienced Clinicians} \cite{neves2021interpretable}, where junior clinicians perceived improvements in risk assessment, (1.2) \textit{Reduction of False Learning} \cite{ellenrieder2023promoting}, where clinicians were critically able to spot erroneous recommendations, and (1.3) \textit{Enhancement of Clinical Planning} \cite{abraham2023integrating}, implying that doctors could better schedule treatments regarding their patients.

However, incorporating XAI may still pose challenges and limitations. 
Therefore, we discuss themes that emerged from studies expressing neutral perceptions of XAI.
Firstly, several works align with \textbf{Theme 2: Preference for Established Clinical Practice} (\cite{kovalchuk2022three,matthiesen2021clinician,zhang2024rethinking,sivaraman2023ignore,hwang2022clinical}). For instance, \cite{matthiesen2021clinician} argued that, although XAI tools were perceived as potentially beneficial for transparency, they did not lead to substantial changes in clinical decision-making. Similarly, \cite{sivaraman2023ignore} observed that XAI outputs were often regarded as supplementary evidence rather than as decisive information. Moreover,~\cite{zhang2024rethinking} noted that it may be more effective to decouple XAI from the final decision stage, positioning it instead as a supporting mechanism during model validation or clinician training. Furthermore, we also identify the sub-them (2.1) \textit{XAI not Always Actionable} (\cite{bhattacharya2023directive,chari2023informing,ihongbe2024evaluating}), where XAI often lacks clinical contextualization, making it insufficiently aligned with established diagnostic reasoning. 
Collectively, these findings suggest that while clinicians acknowledge the theoretical advantages of XAI, their reliance on established diagnostic reasoning and professional experience continues to dominate practical decision processes.

Next, we identified \textbf{Theme 3: Disparities Across Clinical Roles and Seniority Levels}:~\cite{gombolay2024effects,barda2020qualitative,cabitza2025fromoracular}. A consistent observation across these studies was that clinicians’ experiences with CDSS varied according to their level of expertise. For instance,~\cite{cabitza2025fromoracular} reported that less experienced clinicians perceived the CDSS as cognitively demanding. Conversely, ~\cite{gombolay2024effects} found that more senior clinicians encountered greater difficulties integrating the system into their established workflows. Likewise,~\cite{barda2020qualitative} observed that clinical roles significantly shaped attitudes toward CDSS adoption, with preference given to solutions that minimized cognitive load and aligned with existing professional practices. These findings indicate that user acceptance and system usability in XAI-supported clinical environments are not uniform but mediated by clinicians’ roles, expertise, and cognitive expectations.

Finally, we determined \textbf{Theme 4: Frictions Between Stakeholders for Development}:~\cite{Bienefeld2023}. Although only one study fits into this category, it is worth to discuss the implications of this theme. Here,~\cite{Bienefeld2023} documented that frictions may arise between clinicians (as the CDSS users) and developers. Specifically, clinicians emphasized the importance of clinical plausibility and the alignment of model outputs with established medical reasoning, whereas developers prioritized model interpretability. 
Therefore, it may be logical to presume that early aligning stakeholders objectives is necessary to properly build usable CDSS without frictions.
Contextually, we highlight that in only 12 studies (out of 31) early clinicians’ feedback was gathered before CDSS development \cite{barda2020qualitative,matthiesen2021clinician,Bienefeld2023,ellenrieder2023promoting,pumplun2023bringing,hwang2022clinical,zhang2024rethinking,bhattacharya2023directive,abraham2023integrating,gombolay2024effects,jing2025development}. Early incorporation of clinicians’ perspectives can help ensure that system design and model outputs are aligned with real-world clinical expectations. 

Last but not least, however, only nine of these studies were conducted with reasonable sample sizes, \textit{e.g.}, $>$30 participants. This limitation may impact the statistical generalizability of the findings previously discussed.

\begin{table*}
    \centering
    \caption{Summary of discovered themes related to the works surveyed, pointing out clinical outcomes and clinical takeaways.}
    \label{tab:themes_analysis}
    \begin{tabular}{|p{6cm}|p{4cm}|p{4cm}|}
        \hline
        \textbf{Themes and Sub-themes} & \textbf{Papers} & \textbf{Clinical Outcomes} \\
        \hline

        \textbf{1. Improvements in Risk Assessments} & 
        \makecell{
        \cite{brennan2019comparing} \\ 
        \cite{he2024vms} \\ 
        \cite{jung2025evaluating} \\ 
        \cite{hur2025comparison} \\ 
        \cite{kumarakulasinghe2020evaluating} \\ 
        \cite{gu2020acase} \\ 
        \cite{jing2025development} \\ 
        \cite{rajashekar2024human} \\ 
        \cite{famiglini2024evidence} \\ 
        \cite{rainey2024operationalizing} \\ 
        \cite{sabol2020explainable} \\ 
        \cite{kayadibi2025ai} \\ 
        \cite{singh2021evaluation} \\ 
        \cite{chanda2024dermatologist} \\ 
        \cite{pumplun2023bringing} \\ 
        \cite{singla2023medical}
        } 
        & \makecell{Documented Adoption \\ Enhanced Diagnostics} \\

        \hspace{1em}1.1 Enhanced Support for Novice Clinicians & 
        \makecell{\cite{neves2021interpretable}} & 
        \makecell{Better Novice Support} \\

        \hspace{1em}1.2 Reduction of False Learning & 
        \makecell{\cite{ellenrieder2023promoting}} & 
        \makecell{Optimal Learning} \\

        \hspace{1em}1.3 Enhancement of Clinical Planning & 
        \makecell{\cite{abraham2023integrating}} & 
        \makecell{Enhanced Planning} \\
        \hline

        \textbf{2. Preference for Established Clinical Practice} & 
        \makecell{
        \cite{kovalchuk2022three} \\ 
        \cite{matthiesen2021clinician} \\ 
        \cite{zhang2024rethinking} \\ 
        \cite{sivaraman2023ignore} \\ 
        \cite{hwang2022clinical}
        } 
        & \makecell{Seek for Clinical \\ Explanations Plausibility} \\

        \hspace{1em}2.1 XAI not Always Actionable & 
        \makecell{
        \cite{bhattacharya2023directive} \\ 
        \cite{ihongbe2024evaluating} \\ 
        \cite{chari2023informing}
        } 
        & \makecell{Prefer Actionable \\ Explanations}  \\
        \hline

        \textbf{3. Disparities Across Clinical Roles / Seniority Levels} & 
        \makecell{
        \cite{gombolay2024effects} \\ 
        \cite{barda2020qualitative} \\ 
        \cite{cabitza2025fromoracular}
        } 
        & \makecell{Implement Adaptive \\ Explanations } \\
        \hline
        
        \textbf{4. Frictions Between Stakeholders for Development} & 
        \makecell{\cite{Bienefeld2023}} 
        & \makecell{Integrate Iterative \\ Feedback Loops} \\
        \hline
    \end{tabular}
\end{table*}

\subsection{A Socio-Technical Perspective for XAI-based CDSS}

As previously discussed, Theme 1 can positively promote CDSS adoption in healthcare.
However, Themes 2, 3, and 4 may generate frictions regarding CDSS usage. 
Under a theoretical perspective, such frictions may be attributed to socio-technical gaps between human requirements and the effective utility of technological solutions, because human activity is \textit{``flexible and nuanced"}, while computational mechanisms are \textit{``rigid and brittle"}~\cite{ackerman2000intellectual}.
In other words, a socio-technical gap refers to the discrepancy between human needs, including expectations, and the technological capabilities designed to support them, highlighting the limitations of technological systems in fully accommodating the nuances of human interactions. 
This general theoretical construct can be extended to the purpose of XAI explanations in CDSS, as explanation capabilities might not meet rigorous human requirements from precise clinical knowledge (\textit{e.g.}, Theme 2), given that CDSS are dramatically at high stake, as output recommendations may have a significant impact on patient lives.
Simply put, implementing XAI solutions in CDSS might not always bridge the socio-technical gap when compared to systems that do not adopt XAI, or systems that use basic reasoning in current clinical practice. 

Therefore, based on the thematic analysis in Section \ref{sec:thematic_analysis}, we suggest a reference protocol for effective CDSS development. 
This is an \textbf{iterative} process that requires exploring and connecting both the XAI and HCI fields, leveraging their theoretical constructs to eventually devise a practical methodology. 
We begin by identifying two theoretical dimensions: \textbf{T1}: determining stakeholders and their requirements for use case models (\textit{human requirement realism}), and \textbf{T2}: developing XAI evaluation methods to reliably assess whether and to what extent these requirements are met in downstream use cases (\textit{context realism}) \cite{liao2021human}.

Bridging the thematic findings to the two theoretical dimensions, we argue that both \textbf{T1 (human requirement realism)} and \textbf{T2 (context realism)} are interrelated across all four themes previously identified. 
Specifically, \textbf{Theme~1: Improvements in Risk Assessments} relates to T2 through the empirical evaluation of XAI effectiveness in enhancing diagnostic performance, while T1 ensures that such improvements correspond to clinicians' perceived needs in decision support. 
\textbf{Theme~2: Preference for Established Clinical Practice} connects to T2 by assessing whether XAI integration preserves existing diagnostic workflows, and to T1 by accounting for clinicians’ expectations regarding interpretability. 
\textbf{Theme~3: Disparities Across Clinical Roles / Seniority Levels} maps to T1 by identifying different informational requirements across stakeholders, and to T2 by evaluating whether explanation mechanisms adequately support stakeholders. 
Finally, \textbf{Theme~4: Frictions Between Stakeholders for Development} aligns with T1 through early participation to align stakeholders objectives, and with T2 through iterative evaluation of feedback. 

Stakeholder requirements (human requirement realism) must be mapped to the concrete CDSS use case (context realism); the smaller the mismatch, the smaller the socio-technical gap.
%
%
%
Practically,
designing a human-centered CDSS that conforms to such principles involves creating a system that is \textbf{accurate} in treatment recommendations, thus improving risk assessments and clinicians learning (Theme 1), \textbf{explainable} in how and why such recommendations are conveyed, thus seeking for clinically plausible, adaptive and actionable explanations (Theme 1, Theme 2), easily \textbf{usable} by any clinician, thus neither disrupting current clinical practice nor imposing a significant cognitive burden (Theme 2, Theme 3), and \textbf{seamlessly integrable} into existing technological workflows, aligning all stakeholders' objectives, from developers to clinicians, among others (Theme 4).  
Leveraging this knowledge, we propose a CDSS protocol in the next sub-section. 


\subsection{Stakeholder-Centric Protocol for CDSS Development}  

Practically, the development of CDSS requires an iterative, stakeholder-centric methodology that balances clinical expectations with clinical usability. We propose a four-phase framework: (1) stakeholders and their requirements identification, (2) CDSS goal(s) refinement for the contextual use case and informing stakeholders, (3) prototype development, and (4) evaluation. We argue that phases (3) and (4) should be  explicitly iterative, allowing for continuous refinement. 

\subsubsection*{(1) Stakeholder Identification and Requirements}  
Core stakeholders include clinicians, developers, and healthcare organizations (hospitals, clinics, management). However, for real-world adoption, regulators, payers/insurers, and policymakers must also be considered \cite{shortliffe2018clinical}.  
Stakeholder requirements are often divergent, as clinicians may emphasize improvements in risk assessment and XAI non-disruptive actionability (Themes 1,2), while managers may focus on efficiency, and regulators require compliance. This process inevitably necessitates to negotiate trade-offs \cite{vandepoel2013values}.  
Although our review shows that empirical evaluations targeted clinicians, we argue that patients should also be considered as active stakeholders \cite{ancker2021invisible}. Patients are directly affected by AI-informed decisions and increasingly encounter CDSS outputs in decision-making contexts. 
This absence highlights a non-negligible research gap: future work may design evaluation protocols that may incorporate patient perspectives, ensuring that explanations are not only useful by clinicians but also easily communicable to patients (when possible).

\subsubsection*{(2) CDSS Goal(s) Refinement for the Contextual Use Case}  
Stakeholders would collaborate to define the clinical problem the CDSS will address to preventively predict any possible friction later (Theme 4). Shared problem framing ensures that system design aligns with actual decision-making needs rather than abstract performance benchmarks. Non-technical stakeholders must be informed of the CDSS’s underlying logic, including its machine learning models, XAI components, and evaluation frameworks. 
This stage establishes the evaluation criteria and metrics that guide subsequent development. To this end,~\cite{Brankovic2025} formally proposed a reference set of 14 evaluation metrics for XAI in CDSS divided into clinical attributes, decision attributes, and model attributes that all stakeholders should be aware of. 

\subsubsection*{(3) Prototype Development}  
A progressive validation pipeline is recommended: initial testing with low-fidelity prototypes to explore the general structure, followed by retrospective studies in which clinicians may provide early feedback for more rapid iteration, and culminating in high-fidelity prototypes that closely resemble the final product in functionality. 
Prototypes should seamlessly integrate into existing clinical workflows (Theme 2) \cite{weir2021cognitive}. Here, human-centered usability testing, such as think-aloud studies, can capture clinicians’ perceptions of system accuracy and explainability during interaction.  

\subsubsection*{(4) Evaluation and Iteration}  
Evaluation serves both as a quality assurance mechanism and as feedback for redesign. It must address:  
\begin{itemize}  
    \item \textbf{Technical validation}: predictive accuracy, robustness, explainability, fairness, and safety.  
    \item \textbf{Human-centered evaluation}: trust, cognitive workload, workflow integration, and overall usability \cite{khairat2018reasons}.  
    \item \textbf{Ethical-legal considerations}: accountability, transparency, and bias mitigation \cite{ghassemi2021false}.  
\end{itemize}  

Phases (3) and (4) should be conceived as an iterative cycle, in which evaluation outcomes systematically inform redesign. This process should be guided by predefined checkpoints and measurable criteria (\textit{e.g.}, usability thresholds, calibration performance) to ensure that adjustments are specifically targeted.

\subsubsection*{Field-Agnostic but Context-Sensitive}  
Although the framework is field-agnostic, it requires adaptation to local contexts. For instance, radiology-oriented CDSS may emphasize imaging pipelines, whereas psychiatry may require interpretability in narrative or unstructured data. Hence, the framework is best understood as \textit{generalizable but context-sensitive}, consistent with socio-technical perspectives on CDSS design \cite{greenes2014cds}.

\section{Conclusion}

In this work, we examined the role of explainability in CDSS and the socio-technical challenges it faces.
While explainability generally enhances clinicians’ trust in risk assessment, its effectiveness varies with clinicians’ expertise, XAI actionability, and the barriers that current clinical practice presents. 
The socio-technical gap remains a central challenge, as human decision-making is inherently flexible, whereas CDSS operate within rigid constraints. 
Given the high stakes in healthcare, rigorous evaluation strategies, particularly application-grounded methodologies, are essential to align CDSS with stakeholder needs. To address these challenges, we proposed an iterative framework that integrates XAI and HCI principles. By engaging key stakeholders and refining CDSS through continuous evaluation, we advocate for a development approach that prioritizes usability, trust, and clinical integration.  Notably, we argue that rather than replacing human expertise, CDSS should serve as augmentative tools that enhance clinicians’ decision-making. 
By synthesizing evidence on human-centered evaluations of XAI in CDSS and proposing a stakeholder-centric framework, this survey contributes to the advancement of trustworthy and integrative healthcare artificial intelligence, offering a roadmap for the future design and deployment of clinically viable CDSS.

\section*{Acknowledgments}
This work was funded by Fundação para a Ciência e a Tecnologia (PRT/BD/154926/2023, UIDB/00124/2025,
UID/PRR/124/2025, Nova School of Business and Economics, UID/04516/NOVA Laboratory for Computer Science and Informatics (NOVA LINCS), 2024.07361.IACDC, Project HyCARE), and LISBOA2030
(DataLab2030 - LISBOA2030-FEDER-01314200).


\bibliographystyle{named}
\bibliography{ijcai25}

@String{Computing = "Computing" }

@String{Computer = "{IEEE} Computer" }

@String{Academic = "Academic Press" }

@String{Springer = "Springer-Verlag" }

@inproceedings{sivaraman2023ignore,
  title={Ignore, trust, or negotiate: understanding clinician acceptance of AI-based treatment recommendations in health care},
  author={Sivaraman, Venkatesh and Bukowski, Leigh A and Levin, Joel and Kahn, Jeremy M and Perer, Adam},
  booktitle={Proceedings of the 2023 CHI Conference on Human Factors in Computing Systems},
  pages={1--18},
  year={2023}
}

@article{wears2005computer,
  title={Computer technology and clinical work: still waiting for Godot},
  author={Wears, Robert L and Berg, Marc},
  journal={Jama},
  volume={293},
  number={10},
  pages={1261--1263},
  year={2005},
  publisher={American Medical Association}
}

@article{khairat2018reasons,
  title={Reasons for physicians not adopting clinical decision support systems: critical analysis},
  author={Khairat, Saif and Marc, David and Crosby, William and Al Sanousi, Ali and others},
  journal={JMIR Medical Informatics},
  volume={6},
  number={2},
  pages={e8912},
  year={2018},
  publisher={JMIR Publications Inc., Toronto, Canada}
}

@article{kohli2018cad,
  title={Why CAD failed in mammography},
  author={Kohli, Ajay and Jha, Saurabh},
  journal={Journal of the American College Radiology},
  volume={15},
  number={3 Pt B},
  pages={535--537},
  year={2018}
}

@incollection{musen2021clinical,
  title={Clinical decision-support systems},
  author={Musen, Mark A and Middleton, Blackford and Greenes, Robert A},
  booktitle={Biomedical informatics: computer applications in health care and biomedicine},
  pages={795--840},
  year={2021},
  publisher={Springer}
}

@article{horsky2012interface,
  title={Interface design principles for usable decision support: a targeted review of best practices for clinical prescribing interventions},
  author={Horsky, Jan and Schiff, Gordon D and Johnston, Douglas and Mercincavage, Lauren and Bell, Douglas and Middleton, Blackford},
  journal={Journal of Biomedical Informatics},
  volume={45},
  number={6},
  pages={1202--1216},
  year={2012},
  publisher={Elsevier}
}

@article{barda2020qualitative,
  title={A qualitative research framework for the design of user-centered displays of explanations for machine learning model predictions in healthcare},
  author={Barda, Amie J and Horvat, Christopher M and Hochheiser, Harry},
  journal={BMC Medical Informatics and Decision Making},
  volume={20},
  pages={1--16},
  year={2020},
  publisher={Springer}
}

@article{bihorac2019mysurgeryrisk,
  title={MySurgeryRisk: development and validation of a machine-learning risk algorithm for major complications and death after surgery},
  author={Bihorac, Azra and Ozrazgat-Baslanti, Tezcan and Ebadi, Ashkan and Motaei, Amir and Madkour, Mohcine and Pardalos, Panagote M and Lipori, Gloria and Hogan, William R and Efron, Philip A and Moore, Frederick and others},
  journal={Annals of Surgery},
  volume={269},
  number={4},
  pages={652--662},
  year={2019},
  publisher={LWW}
}

@article{brennan2019comparing,
  title={Comparing clinical judgment with the MySurgeryRisk algorithm for preoperative risk assessment: a pilot usability study},
  author={Brennan, Meghan and Puri, Sahil and Ozrazgat-Baslanti, Tezcan and Feng, Zheng and Ruppert, Matthew and Hashemighouchani, Haleh and Momcilovic, Petar and Li, Xiaolin and Wang, Daisy Zhe and Bihorac, Azra},
  journal={Surgery},
  volume={165},
  number={5},
  pages={1035--1045},
  year={2019},
  publisher={Elsevier}
}

@inproceedings{sendak2020human,
  title={" The human body is a black box" supporting clinical decision-making with deep learning},
  author={Sendak, Mark and Elish, Madeleine Clare and Gao, Michael and Futoma, Joseph and Ratliff, William and Nichols, Marshall and Bedoya, Armando and Balu, Suresh and O'Brien, Cara},
  booktitle={Proceedings of the 2020 Conference on Fairness, Accountability, and Transparency},
  pages={99--109},
  year={2020}
}

@article{matthiesen2021clinician,
  title={Clinician preimplementation perspectives of a decision-support tool for the prediction of cardiac arrhythmia based on machine learning: near-live feasibility and qualitative study},
  author={Matthiesen, Stina and Diederichsen, S{\o}ren Z{\"o}ga and Hansen, Mikkel Klitzing Hartmann and Villumsen, Christina  et al.
                                          
         },
  journal={JMIR Human Factors},
  volume={8},
  number={4},
  pages={e26964},
  year={2021},
  publisher={JMIR Publications Inc., Toronto, Canada}
}

@inproceedings{ellenrieder2023promoting,
  title={Promoting Learning Through Explainable Artificial Intelligence: An
Experimental Study in RadiologyExperimental Study in Radiology},
  author={Ellenrieder, Sara and Kallina, Emma Marlene and Pumplun, Luisa and Gawlitza, Joshua Felix and Ziegelmayer, Sebastian and Buxmann, Pete},
  booktitle={Proceedings of International Conference on Information Systems},
  pages={1-17},
  year={2023}
}

@article{hwang2022clinical,
  title={A clinical decision support system for sleep staging tasks with explanations from artificial intelligence: user-centered design and evaluation study},
  author={Hwang, Jeonghwan and Lee, Taeheon and Lee, Honggu and Byun, Seonjeong},
  journal={Journal of Medical Internet Research},
  volume={24},
  number={1},
  pages={e28659},
  year={2022},
  publisher={JMIR Publications Toronto, Canada}
}

@inproceedings{kumarakulasinghe2020evaluating,
  title={Evaluating local interpretable model-agnostic explanations on clinical machine learning classification models},
  author={Kumar., Nesaretnam Barr and Blomberg, Tobias and Liu, Jintai and Leao, Alexandra Saraiva and Papapetrou, Panagiotis},
  booktitle={2020 IEEE 33rd International Symposium on Computer-Based Medical Systems (CBMS)},
  pages={7--12},
  year={2020},
  organization={IEEE}
}

@article{kovalchuk2022three,
  title={Three-stage intelligent support of clinical decision making for higher trust, validity, and explainability},
  author={Kovalchuk, Sergey V and Kopanitsa, Georgy D and Derevitskii, Ilia V and Matveev, Georgy A and Savitskaya, Daria A},
  journal={Journal of Biomedical Informatics},
  volume={127},
  pages={104013},
  year={2022},
  publisher={Elsevier}
}

@article{sabol2020explainable,
  title={Explainable classifier for improving the accountability in decision-making for colorectal cancer diagnosis from histopathological images},
  author={Sabol, Patrik and Sin{\v{c}}{\'a}k, Peter and Hartono, Pitoyo and Ko{\v{c}}an, Pavel and Benetinov{\'a}, Zuzana and Blich{\'a}rov{\'a}, Al{\v{z}}beta and Verb{\'o}ov{\'a}, L'udmila and {\v{S}}tammov{\'a}, Erika and Sabolov{\'a}-Fabianov{\'a}, Ant{\'o}nia and Ja{\v{s}}kov{\'a}, Anna},
  journal={Journal of Biomedical Informatics},
  volume={109},
  pages={103523},
  year={2020},
  publisher={Elsevier}
}

@inproceedings{ribeiro2016should,
  title={Why should I trust you? Explaining the predictions of any classifier},
  author={Ribeiro, Marco Tulio and Singh, Sameer and Guestrin, Carlos},
  booktitle={Proceedings of the 22nd ACM SIGKDD International Conference on Knowledge Discovery and Data Mining},
  pages={1135--1144},
  year={2016}
}

@inproceedings{rajashekar2024human,
  title={Human-Algorithmic Interaction Using a Large Language Model-Augmented Artificial Intelligence Clinical Decision Support System},
  author={Rajashekar, Niroop Channa and Shin, Yeo Eun and Pu, Yuan and Chung, Sunny and You, Kisung and Giuffre, Mauro and Chan, Colleen E and Saarinen, Theo and Hsiao, Allen and Sekhon, Jasjeet and others},
  booktitle={Proceedings of the CHI Conference on Human Factors in Computing Systems},
  pages={1--20},
  year={2024}
}

@book{molnar2022,
  title      = {Interpretable Machine Learning},
  author     = {Christoph Molnar},
  year       = {2022},
  subtitle   = {A Guide for Making Black Box Models Explainable},
  edition    = {2},
  url        = {https://christophm.github.io/interpretable-ml-book}
}

@inproceedings{ribeiro2018anchors,
  title={Anchors: High-precision model-agnostic explanations},
  author={Ribeiro, Marco Tulio and Singh, Sameer and Guestrin, Carlos},
  booktitle={Proceedings of the AAAI conference on artificial intelligence},
  volume={32},
  number={1},
  year={2018}
}

@article{friedman2001greedy,
  title={Greedy function approximation: a gradient boosting machine},
  author={Friedman, Jerome H},
  journal={Annals of statistics},
  pages={1189--1232},
  year={2001},
  publisher={JSTOR}
}

@article{apley2020visualizing,
  title={Visualizing the effects of predictor variables in black box supervised learning models},
  author={Apley, Daniel W and Zhu, Jingyu},
  journal={Journal of the Royal Statistical Society Series B: Statistical Methodology},
  volume={82},
  number={4},
  pages={1059--1086},
  year={2020},
  publisher={Oxford University Press}
}

@article{doshi2017towards,
  title={Towards a rigorous science of interpretable machine learning},
  author={Doshi-Velez, Finale and Kim, Been},
  journal={arXiv preprint arXiv:1702.08608},
  year={2017}
}

@article{liao2021human,
  title={Human-centered explainable ai (xai): From algorithms to user experiences},
  author={Liao, Q Vera and Varshney, Kush R},
  journal={arXiv preprint arXiv:2110.10790},
  year={2021}
}

@article{ledley1959reasoning,
  title={Reasoning foundations of medical diagnosis: symbolic logic, probability, and value theory aid our understanding of how physicians reason},
  author={Ledley, Robert S and Lusted, Lee B},
  journal={Science},
  volume={130},
  number={3366},
  pages={9--21},
  year={1959},
  publisher={American Association for the Advancement of Science}
}

@inproceedings{selvaraju2017grad,
  title={Grad-cam: Visual explanations from deep networks via gradient-based localization},
  author={Selvaraju, Ramprasaath R and Cogswell, Michael and Das, Abhishek and Vedantam, Ramakrishna and Parikh, Devi and Batra, Dhruv},
  booktitle={Proceedings of the IEEE International Conference on Computer Vision},
  pages={618--626},
  year={2017}
}

@article{pumplun2023bringing,
  title={Bringing machine learning systems into clinical practice: a design science approach to explainable machine learning-based clinical decision support systems},
  author={Pumplun, Luisa and Peters, Felix and Gawlitza, Joshua F and Buxmann, Peter},
  journal={Journal of the Association for Information Systems},
  volume={24},
  number={4},
  pages={953--979},
  year={2023}
}

@inproceedings{sundararajan2017axiomatic,
  title={Axiomatic attribution for deep networks},
  author={Sundararajan, Mukund and Taly, Ankur and Yan, Qiqi},
  booktitle={International Conference on Machine Learning},
  pages={3319--3328},
  year={2017},
  organization={PMLR}
}

@inproceedings{zhang2024rethinking,
  title={Rethinking human-ai collaboration in complex medical decision making: A case study in sepsis diagnosis},
  author={Zhang, Shao and Yu, Jianing and Xu, Xuhai and Yin, Changchang and Lu, Yuxuan and Yao, Bingsheng and Tory, Melanie and Padilla, Lace M and Caterino, Jeffrey and Zhang, Ping and others},
  booktitle={Proceedings of the CHI Conference on Human Factors in Computing Systems},
  pages={1--18},
  year={2024}
}

@article{singh2021evaluation,
  title={Evaluation of explainable deep learning methods for ophthalmic diagnosis},
  author={Singh, Amitojdeep and Jothi Balaji, Janarthanam and Rasheed, Mohammed Abdul and Jayakumar, Varadharajan and Raman, Rajiv and Lakshminarayanan, Vasudevan},
  journal={Clinical Ophthalmology},
  pages={2573--2581},
  year={2021},
  publisher={Taylor \& Francis}
}

@article{neves2021interpretable,
  title={Interpretable heartbeat classification using local model-agnostic explanations on ECGs},
  author={Neves, In{\^e}s and Folgado, Duarte and Santos, Sara and Barandas, Marilia and Campagner, Andrea and Ronzio, Luca and Cabitza, Federico and Gamboa, Hugo},
  journal={Computers in Biology and Medicine},
  volume={133},
  pages={104393},
  year={2021},
  publisher={Elsevier}
}

@article{loh2022application,
  title={Application of explainable artificial intelligence for healthcare: A systematic review of the last decade (2011--2022)},
  author={Loh, Hui Wen and Ooi, Chui Ping and Seoni, Silvia and Barua, Prabal Datta and Molinari, Filippo and Acharya, U Rajendra},
  journal={Computer Methods and Programs in Biomedicine},
  volume={226},
  pages={107161},
  year={2022},
  publisher={Elsevier}
}

@inproceedings{speith2022review,
  title={A review of taxonomies of explainable artificial intelligence (XAI) methods},
  author={Speith, Timo},
  booktitle={Proceedings of the 2022 ACM Conference on Fairness, Accountability, and Transparency},
  pages={2239--2250},
  year={2022}
}

@inproceedings{lundberg2017shap,
title = {A Unified Approach to Interpreting Model Predictions},
author = {Lundberg, Scott M and Lee, Su-In},
booktitle = {Advances in Neural Information Processing Systems 30},
editor = {I. Guyon and U. V. Luxburg and S. Bengio and H. Wallach and R. Fergus and S. Vishwanathan and R. Garnett},
pages = {4765--4774},
year = {2017},
publisher = {Curran Associates, Inc.},
url = {http://papers.nips.cc/paper/7062-a-unified-approach-to-interpreting-model-predictions.pdf}
}

@article{montavon2017explaining,
  title={Explaining nonlinear classification decisions with deep taylor decomposition},
  author={Montavon, Gr{\'e}goire and Lapuschkin, Sebastian and Binder, Alexander and Samek, Wojciech and M{\"u}ller, Klaus-Robert},
  journal={Pattern Recognition},
  volume={65},
  pages={211--222},
  year={2017},
  publisher={Elsevier}
}

@article{ackerman2000intellectual,
  title={{The intellectual challenge of CSCW: the gap between social requirements and technical feasibility}},
  author={Ackerman, Mark S},
  journal={Human--Computer Interaction},
  volume={15},
  number={2-3},
  pages={179--203},
  year={2000},
  publisher={Taylor \& Francis}
}

@article{liao2023rethinking,
  title={Rethinking model evaluation as narrowing the socio-technical gap},
  author={Liao, Q Vera and Xiao, Ziang},
  journal={arXiv preprint arXiv:2306.03100},
  year={2023}
}

@techreport{sahni2023potential,
  title={The potential impact of artificial intelligence on healthcare spending},
  author={Sahni, Nikhil and Stein, George and Zemmel, Rodney and Cutler, David M},
  year={2023},
  institution={National Bureau of Economic Research Cambridge, MA, USA:}
}

@techreport{goldsack2024healthcareai,
    title = {Billions of Dollars Have Been Invested in Healthcare AI. But Are We Spending in the Right Places?},
    author = {Goldsack, Jennifer and Overgaard, Shauna},
    institution = {World Economic Forum},
    year = {2024}
}

@article{zhang2023ethics,
  title={Ethics and governance of trustworthy medical artificial intelligence},
  author={Zhang, Jie and Zhang, Zong-ming},
  journal={BMC Medical Informatics and Decision Making},
  volume={23},
  number={1},
  pages={7},
  year={2023},
  publisher={Springer}
}

@article{ghassemi2021false,
  title={The false hope of current approaches to explainable artificial intelligence in health care},
  author={Ghassemi, Marzyeh and Oakden-Rayner, Luke and Beam, Andrew L},
  journal={The Lancet Digital Health},
  volume={3},
  number={11},
  pages={e745--e750},
  year={2021},
  publisher={Elsevier}
}

@article{amann2022explain,
  title={{To explain or not to explain?—Artificial intelligence explainability in clinical decision support systems}},
  author={Amann, Julia and Vetter, Dennis and Blomberg, Stig Nikolaj and Christensen, Helle Collatz and Coffee, Megan and Gerke, Sara and Gilbert, Thomas K and Hagendorff, Thilo and Holm, Sune and Livne, Michelle and others},
  journal={PLOS Digital Health},
  volume={1},
  number={2},
  pages={e0000016},
  year={2022},
  publisher={Public Library of Science San Francisco, CA USA}
}

@article{lorenzini2023artificial,
  title={Artificial intelligence and the doctor--patient relationship expanding the paradigm of shared decision making},
  author={Lorenzini, Giorgia and others},
  journal={Bioethics},
  volume={37},
  number={5},
  pages={424--429},
  year={2023}
}

@article{qin2024personalization,
  title={Examining the impact of personalization and carefulness in AI-generated health advice: Trust, adoption, and insights in online healthcare consultations experiments},
  author={Qin, Hongyi and others},
  journal={Technology in Society},
  volume={79},
  pages={102726},
  year={2024}
}

@article{longoni2019resistance,
  title={Resistance to medical artificial intelligence},
  author={Longoni, Chiara and Bonezzi, Andrea and Morewedge, Carey K.},
  journal={Journal of Consumer Research},
  volume={46},
  number={4},
  pages={629--650},
  year={2019}
}

@article{gaczek2023overcoming,
  title={Overcoming consumer resistance to AI in general health care},
  author={Gaczek, Piotr and others},
  journal={Journal of Interactive Marketing},
  volume={58},
  number={2-3},
  pages={321--338},
  year={2023}
}

@article{liu2021need,
  title={Need for intensive care for children with pneumonia},
  author={Liu, Hui-Wen and others},
  journal={Computer Methods and Programs in Biomedicine},
  volume={226},
  pages={107161},
  year={2022}
}

@article{shi2022explainable,
  title={Explainable machine learning model for predicting the occurrence of postoperative malnutrition in children with congenital heart disease},
  author={Shi, H and others},
  journal={Clinical Nutrition},
  volume={41},
  number={1},
  pages={202--210},
  year={2022}
}

@article{chen2021forecasting,
  title={Forecasting adverse surgical events using self-supervised transfer learning for physiological signals},
  author={Chen, Hugh and Lundberg, Scott M and Erion, Gabriel and Kim, Jerry H and Lee, Su-In},
  journal={NPJ Digital Medicine},
  volume={4},
  number={1},
  pages={167},
  year={2021}
}

@article{nguyen2021prediction,
  title={Budget constrained machine learning for early prediction of adverse outcomes for COVID-19 patients},
  author={Nguyen, Sa and others},
  journal={Scientific Reports},
  volume={11},
  number={1},
  pages={19543},
  year={2021}
}

@article{zeng2021explainable,
  title={Explainable machine-learning predictions for complications after pediatric congenital heart surgery},
  author={Zeng, Xinlong and others},
  journal={Scientific Reports},
  volume={11},
  number={1},
  pages={17244},
  year={2021}
}

@article{zhang2021explainable,
  title={An explainable supervised machine learning predictor of acute kidney injury after adult deceased donor liver transplantation},
  author={Zhang, Yong and others},
  journal={Journal of Translational Medicine},
  volume={19},
  number={1},
  pages={321},
  year={2021}
}

@article{antoniadi2021prediction,
  title={Prediction of caregiver quality of life in amyotrophic lateral sclerosis using explainable machine learning},
  author={Antoniadi, Anna M and Galvin, Mark and Heverin, Mark and Hardiman, Orla and Mooney, Catherine},
  journal={Scientific Reports},
  volume={11},
  number={1},
  pages={12237},
  year={2021}
}

@article{figueroa2022interpretable,
  title={Interpretable deep learning approach for oral cancer classification using guided attention inference network},
  author={Figueroa, Karla C and others},
  journal={Journal of Biomedical Optics},
  volume={27},
  number={1},
  year={2022}
}

@article{xu2021clinical,
  title={The clinical value of explainable deep learning for diagnosing fungal keratitis using in vivo confocal microscopy images},
  author={Xu, Feng and others},
  journal={Frontiers in Medicine},
  volume={8},
  year={2021}
}

@article{chetoui2021explainable,
  title={Explainable COVID-19 detection on chest X-rays using an end-to-end deep convolutional neural network architecture},
  author={Chetoui, Marwen and Akhloufi, Moulay A and Yousefi, Bardia and Bouattane, El Mustapha},
  journal={Big Data and Cognitive Computing},
  volume={5},
  number={4},
  pages={73},
  year={2021}
}

@article{shi2021covid,
  title={COVID-19 automatic diagnosis with radiographic imaging: explainable attention transfer deep neural networks},
  author={Shi, Weiming and Tong, Ling and Zhu, Yingjuan and Wang, May D},
  journal={IEEE Journal of Biomedical and Health Informatics},
  volume={25},
  number={7},
  pages={2376--2387},
  year={2021}
}

@article{ozturk2020automated,
  title={Automated detection of COVID-19 cases using deep neural networks with X-ray images},
  author={Ozturk, Tulin and Talo, Muhammed and Yildirim, Eylul Azra and Baloglu, Ulas Baran and Yildirim, Ozal and Acharya, U Rajendra},
  journal={Computers in Biology and Medicine},
  volume={121},
  pages={103792},
  year={2020}
}

@article{yoo2021xecgnet,
  title={xECGNet: fine-tuning attention map within convolutional neural network to improve detection and explainability of concurrent cardiac arrhythmias},
  author={Yoo, Jaebum and Jun, Tae Joon and Kim, Young-Hak},
  journal={Computer Methods and Programs in Biomedicine},
  volume={208},
  pages={106281},
  year={2021}
}

@article{thakoor2021robust,
  title={Robust and interpretable convolutional neural networks to detect glaucoma in optical coherence tomography images},
  author={Thakoor, Kiran A and Koorathota, Sunil C and Hood, Donald C and Sajda, Paul},
  journal={IEEE Transactions on Biomedical Engineering},
  volume={68},
  number={8},
  pages={2456--2466},
  year={2021}
}

@article{binder2021morphological,
  title={Morphological and molecular breast cancer profiling through explainable machine learning},
  author={Binder, Alexander and others},
  journal={Nature Machine Intelligence},
  volume={3},
  pages={355--366},
  year={2021}
}

@article{dong2021explainable,
  title={Explainable automated coding of clinical notes using hierarchical label-wise attention networks and label embedding initialisation},
  author={Dong, Hang and Su{\'a}rez-Paniagua, V{\'\i}ctor and Whiteley, William and Wu, Honghan},
  journal={Journal of Biomedical Informatics},
  volume={116},
  pages={103728},
  year={2021}
}

@article{hu2021explainable,
  title={An explainable CNN approach for medical codes prediction from clinical text},
  author={Hu, Shanshan and Teng, Fei and Huang, Lin and Yan, Jun and Zhang, Honglong},
  journal={BMC Medical Informatics and Decision Making},
  volume={21},
  number={S9},
  pages={256},
  year={2021}
}

@article{gunning2019darpa,
  title={DARPA's Explainable Artificial Intelligence (XAI) Program},
  author={Gunning, David and Aha, David W},
  journal={AI Magazine},
  volume={40},
  number={2},
  pages={44--58},
  year={2019},
  publisher={AAAI Press},
  doi={10.1609/aimag.v40i2.2850}
}

@article{Brankovic2025,
  author       = {Brankovic, A. M. and others},
  title        = {CLIX-M: Clinician-informed XAI evaluation checklist},
  journal      = {npj Digital Medicine},
  year         = {2025},
  volume       = {8},
  pages        = {364}
}

@article{Bienefeld2023,
  author       = {Bienefeld, N. and Boss, J. M. and Lüthy, R. and Brodbeck, D. and Azzati, J. and Blaser, M. and others},
  title        = {Solving the explainable AI conundrum by bridging clinicians’ needs and developers’ goals},
  journal      = {npj Digital Medicine},
  year         = {2023},
  volume       = {6},
  pages        = {94},
  doi          = {10.1038/s41746-023-00837-4}
}

@article{Kim2024,
  author       = {Kim, Y. and others},
  title        = {Human-centered evaluation of XAI applications: a systematic review},
  journal      = {Frontiers in Artificial Intelligence},
  year         = {2024},
  volume       = {7},
  pages        = {1456486}
}

@article{shortliffe2018clinical,
  title={Clinical decision support in the era of artificial intelligence},
  author={Shortliffe, Edward H and Sep{\'u}lveda, Martin J},
  journal={JAMA},
  volume={320},
  number={21},
  pages={2199--2200},
  year={2018},
  publisher={American Medical Association}
}

@article{ancker2021invisible,
  title={The invisible work of personal health information management among people with multiple chronic conditions: qualitative interview study among patients and providers},
  author={Ancker, Jessica S and Witteman, Holly O and Hafeez, Bilal and Provencher, Tania and Van de Velde, Sarah and Carrington, Miranda},
  journal={Journal of Medical Internet Research},
  volume={23},
  number={6},
  pages={e26112},
  year={2021},
  publisher={JMIR Publications}
}

@incollection{vandepoel2013values,
  title={Translating values into design requirements},
  author={van de Poel, Ibo},
  booktitle={Philosophy and Engineering: Reflections on Practice, Principles and Process},
  pages={253--266},
  year={2013},
  publisher={Springer}
}

@article{weir2021cognitive,
  title={A cognitive task analysis of information management strategies in a computerized provider order entry environment},
  author={Weir, Charlene R and Nebeker, Jonathan R and Hicken, Bret L and Campo, Reinaldo and Drews, Frank and Lebar, Brad},
  journal={Journal of the American Medical Informatics Association},
  volume={28},
  number={1},
  pages={30--39},
  year={2021},
  publisher={Oxford University Press}
}

@book{greenes2014cds,
  title={Clinical Decision Support: The Road Ahead},
  author={Greenes, Robert A},
  year={2014},
  publisher={Academic Press}
}

@inproceedings{bhattacharya2023directive,
 address = {New York, NY, USA},
 author = {Bhattacharya, Aditya and Ooge, Jeroen and Stiglic, Gregor and Verbert, Katrien},
 booktitle = {Proceedings of the 28th International Conference on Intelligent User Interfaces},
 doi = {10.1145/3581641.3584075},
 isbn = {9798400701061},
 location = {Sydney, NSW, Australia},
 numpages = {16},
 pages = {204–219},
 publisher = {Association for Computing Machinery},
 series = {IUI '23},
 title = {Directive Explanations for Monitoring the Risk of Diabetes Onset: Introducing Directive Data-Centric Explanations and Combinations to Support What-If Explorations},
 url = {https://doi.org/10.1145/3581641.3584075},
 year = {2023}
}

@article{abraham2023integrating,
 article-number = {104270},
 author = {Abraham, Joanna and Bartek, Brian and Meng, Alicia and King, Christopher
Ryan and Xue, Bing and Lu, Chenyang and Avidan, Michael S.},
 doi = {10.1016/j.jbi.2022.104270},
 earlyaccessdate = {DEC 2022},
 eissn = {1532-0480},
 issn = {1532-0464},
 journal = {JOURNAL OF BIOMEDICAL INFORMATICS},
 month = {JAN},
 researcherid-numbers = {Xue, Bing/IQV-2162-2023
Avidan, Alon/AAK-8865-2020},
 title = {Integrating machine learning predictions for perioperative risk
management: Towards an empirical design of a flexible-standardized risk
assessment tool},
 unique-id = {WOS:000903751900002},
 volume = {137},
 year = {2023}
}

@inproceedings{cabitza2025fromoracular,
 address = {New York, NY, USA},
 author = {Cabitza, Federico and Famiglini, Lorenzo and Fregosi, Caterina and Pe, Samuele and Parimbelli, Enea and La Maida, Giovanni Andrea and Gallazzi, Enrico},
 booktitle = {Proceedings of the 30th International Conference on Intelligent User Interfaces},
 doi = {10.1145/3708359.3712157},
 isbn = {9798400713064},
 location = {},
 numpages = {10},
 pages = {745–754},
 publisher = {Association for Computing Machinery},
 series = {IUI '25},
 title = {From Oracular to Judicial: Enhancing Clinical Decision Making through Contrasting Explanations and a Novel Interaction Protocol},
 url = {https://doi.org/10.1145/3708359.3712157},
 year = {2025}
}

@article{chanda2024dermatologist,
 article-number = {524},
 author = {Chanda, Tirtha and Hauser, Katja and Hobelsberger, Sarah and Bucher,
Tabea-Clara and Garcia, Carina Nogueira and Wies, Christoph and Kittler,
Harald and Tschandl, Philipp and Navarrete-Dechent, Cristian and
Podlipnik, Sebastian and Chousakos, Emmanouil and Crnaric, Iva and
Majstorovic, Jovana and Alhajwan, Linda and Foreman, Tanya and Peternel,
Sandra and Sarap, Sergei and Ozdemir, Irem and Barnhill, Raymond L. and
Llamas-Velasco, Mar and Poch, Gabriela and Korsing, Soeren and
Sondermann, Wiebke and Gellrich, Frank Friedrich and Heppt, Markus V.
and Erdmann, Michael and Haferkamp, Sebastian and Drexler, Konstantin
and Goebeler, Matthias and Schilling, Bastian and Utikal, Jochen S. and
Ghoreschi, Kamran and Froehling, Stefan and Krieghoff-Henning, Eva and
Brinker, Titus J. and Reader Study Consortium},
 doi = {10.1038/s41467-023-43095-4},
 eissn = {2041-1723},
 journal = {NATURE COMMUNICATIONS},
 month = {JAN 15},
 number = {1},
 orcid-numbers = {Peralta, Rosario/0009-0007-6014-9877
Chousakos, Emmanouil/0000-0002-8127-4761
Damevska, Katerina/0000-0003-4745-3747
Dragolov, Miroslav/0009-0008-6722-7249
Racz, Emoke/0000-0001-5119-6451
Peternel, Sandra/0000-0001-8590-0451
Wies, Christoph/0000-0001-7136-298X
Poch, Gabriela/0000-0003-3948-2505
Natlitz, geb. Knuver, Jana/0000-0003-1066-6431
Haferkamp, Sebastian/0000-0002-3894-8345
Theofilogiannakou, Panagiota/0009-0005-5815-7726
Bucher, Tabea-Clara/0000-0002-8157-5513
Simeonovski, Viktor/0000-0003-2956-0928
Schuh, Sandra/0000-0002-1470-7619
Salava, Alexander/0000-0001-5471-5894
Ghoreschi, Kamran/0000-0002-5526-7517
Hobelsberger, Sarah/0000-0001-5703-324X
Podlipnik, Sebastian/0000-0003-4150-0522
Ammar, Amr/0009-0006-6475-7557
Welponer, Tobias/0000-0003-3159-0720
Chanda, Tirtha/0009-0009-0649-9301
Kolios, Antonios/0000-0002-3897-4578
KITTLER, HARALD/0000-0002-0051-8016
Nogueira Garcia, Carina/0000-0001-9092-5952
Gellrich, Frank Friedrich/0000-0002-2164-4644
Bondare-Ansberga, Vanda/0000-0003-0939-8003},
 researcherid-numbers = {Damevska, Katerina/G-6044-2019
Peternel, Sandra/O-2506-2018
Haferkamp, Sebastian/ABD-7390-2021
Natlitz, geb. Knuver, Jana/KXR-3938-2024
Sondermann, Wiebke/ACA-9208-2022
Llamas-Velasco, Mar/AFG-6527-2022
Kittler, Harald/AAK-1502-2020
Ozdemir, Irem/OJU-4020-2025
Goebeler, Matthias/ISU-7266-2023
Podlipnik, Sebastian/U-3941-2019
Navarrete-Dechent, Cristian/AEX-9897-2022
},
 title = {Dermatologist-like explainable AI enhances trust and confidence in
diagnosing melanoma},
 unique-id = {WOS:001143918100017},
 volume = {15},
 year = {2024}
}

@article{chari2023informing,
 article-number = {102498},
 author = {Chari, Shruthi and Acharya, Prasant and Gruen, Daniel M. and Zhang,
Olivia and Eyigoz, Elif K. and Ghalwash, Mohamed and Seneviratne, Oshani
and Saiz, Fernando Suarez and Meyer, Pablo and Chakraborty, Prithwish
and McGuinness, Deborah L.},
 doi = {10.1016/j.artmed.2023.102498},
 earlyaccessdate = {FEB 2023},
 eissn = {1873-2860},
 issn = {0933-3657},
 journal = {ARTIFICIAL INTELLIGENCE IN MEDICINE},
 month = {MAR},
 orcid-numbers = {Suarez Saiz, Fernando J/0000-0001-8176-5184
Chari, Shruthi/0000-0003-2946-7870
McGuinness, Deborah/0000-0001-7037-4567},
 researcherid-numbers = {Ghalwash, Mohamed/F-8926-2014
McGuinness, Deborah/HGC-4351-2022
Chari, Shruthi/JXL-9579-2024
Saiz, Fernando/K-8201-2019
},
 title = {Informing clinical assessment by contextualizing post-hoc explanations
of risk prediction models in type-2 diabetes},
 unique-id = {WOS:000933256300001},
 volume = {137},
 year = {2023}
}

@article{famiglini2024evidence,
 article-number = {108042},
 author = {Famiglini, Lorenzo and Campagner, Andrea and Barandas, Marilia and
Maida, Giovanni Andrea La and Gallazzi, Enrico and Cabitza, Federico},
 doi = {10.1016/j.compbiomed.2024.108042},
 earlyaccessdate = {FEB 2024},
 eissn = {1879-0534},
 issn = {0010-4825},
 journal = {COMPUTERS IN BIOLOGY AND MEDICINE},
 month = {MAR},
 orcid-numbers = {famiglini, lorenzo/0000-0002-1934-5899
Gallazzi, Enrico/0000-0001-9287-9937
Barandas, Marilia/0000-0002-9445-4809
La Maida, Giovanni Andrea/0000-0001-5360-1262
},
 researcherid-numbers = {Campagner, Andrea/AAB-4238-2020
Barandas, Marilia/J-3482-2019
Cabitza, Federico/JCO-4001-2023
Gallazzi, Enrico/AFL-6149-2022
Famiglini, Lorenzo/AIA-5978-2022},
 title = {Evidence-based XAI: An empirical approach to design more effective and
explainable decision support systems},
 unique-id = {WOS:001178488700001},
 volume = {170},
 year = {2024}
}

@article{gombolay2024effects,
 author = {Gombolay, Grace Y. and Silva, Andrew and Schrum, Mariah and Gopalan,
Nakul and Hallman-Cooper, Jamika and Dutt, Monideep and Gombolay,
Matthew},
 doi = {10.1002/acn3.52036},
 earlyaccessdate = {APR 2024},
 issn = {2328-9503},
 journal = {ANNALS OF CLINICAL AND TRANSLATIONAL NEUROLOGY},
 month = {MAY},
 number = {5},
 orcid-numbers = {Dutt, Monideep/0000-0003-3727-9255
Gombolay, Grace/0000-0003-4830-7792},
 pages = {1224-1235},
 researcherid-numbers = {Gombolay, Grace/GXH-9077-2022},
 title = {Effects of explainable artificial intelligence in neurology decision
support},
 unique-id = {WOS:001204488500001},
 volume = {11},
 year = {2024}
}

@article{gu2020acase,
 article-number = {101858},
 author = {Gu, Dongxiao and Su, Kaixiang and Zhao, Huimin},
 doi = {10.1016/j.artmed.2020.101858},
 eissn = {1873-2860},
 issn = {0933-3657},
 journal = {ARTIFICIAL INTELLIGENCE IN MEDICINE},
 month = {JUL},
 orcid-numbers = {su, kaixiang/0000-0003-2731-511X
Gu, Dongxiao/0000-0003-3557-009X},
 title = {A case-based ensemble learning system for explainable breast cancer
recurrence prediction},
 unique-id = {WOS:000564615100004},
 volume = {107},
 year = {2020}
}

@inproceedings{he2024vms,
author = {He, Chen and Raj, Vishnu and Moen, Hans and Gr\"{o}hn, Tommi and Wang, Chen and Peltonen, Laura-Maria and Koivusalo, Saila and Marttinen, Pekka and Jacucci, Giulio},
title = {VMS: Interactive Visualization to Support the Sensemaking and Selection of Predictive Models},
year = {2024},
isbn = {9798400705083},
publisher = {Association for Computing Machinery},
address = {New York, NY, USA},
url = {https://doi.org/10.1145/3640543.3645151},
doi = {10.1145/3640543.3645151},
booktitle = {Proceedings of the 29th International Conference on Intelligent User Interfaces},
pages = {229–244},
numpages = {16},
location = {Greenville, SC, USA},
series = {IUI '24'}
}

@article{hur2025comparison,
 article-number = {578},
 author = {Hur, Sujeong and Lee, Yura and Park, Joongheum and Jeon, Yeong Jeong and
Cho, Jong Ho and Cho, Duck and Lim, Dobin and Hwang, Wonil and Cha, Won
Chul and Yoo, Junsang},
 doi = {10.1038/s41746-025-01958-8},
 issn = {2398-6352},
 journal = {NPJ DIGITAL MEDICINE},
 month = {SEP 26},
 number = {1},
 title = {Comparison of SHAP and clinician friendly explanations reveals effects
on clinical decision behaviour},
 unique-id = {WOS:001581709900001},
 volume = {8},
 year = {2025}
}

@article{ihongbe2024evaluating,
 article-number = {e0308758},
 author = {E. Ihongbe, Izegbua and Fouad, Shereen and F. Mahmoud, Taha and
Rajasekaran, Arvind and Bhatia, Bahadar},
 doi = {10.1371/journal.pone.0308758},
 eissn = {1932-6203},
 journal = {PLOS ONE},
 month = {OCT 9},
 number = {10},
 orcid-numbers = {Ihongbe, Izegbua E./0009-0003-3630-5404
Fouad, Shereen/0000-0002-4965-7017},
 title = {Evaluating Explainable Artificial Intelligence (XAI) techniques in chest
radiology imaging through a human-centered Lens},
 unique-id = {WOS:001336858900035},
 volume = {19},
 year = {2024}
}

@article{jing2025development,
 article-number = {e095460},
 author = {Jing, Li and Hua, Peng and Zeng, Shumei and Peng, Qing and Wu, Weizi and
Lv, Luofang and Yue, Liqing and zhong, Hu Jian and Huang, Weihong},
 doi = {10.1136/bmjopen-2024-095460},
 issn = {2044-6055},
 journal = {BMJ OPEN},
 month = {JUL 4},
 number = {7},
 researcherid-numbers = {Wu, Weizi/LMO-2067-2024},
 title = {Development of an interpretable machine learning model for frailty risk
prediction in older adult care institutions: a mixed-methods,
cross-sectional study in China},
 unique-id = {WOS:001522786800001},
 volume = {15},
 year = {2025}
}

@article{jung2025evaluating,
 article-number = {105943},
 author = {Jung, Jinsun and Kang, Sunghoon and Choi, Jeeyae and El-Kareh, Robert
and Lee, Hyungbok and Kim, Hyeoneui},
 doi = {10.1016/j.ijmedinf.2025.105943},
 earlyaccessdate = {MAY 2025},
 eissn = {1872-8243},
 issn = {1386-5056},
 journal = {INTERNATIONAL JOURNAL OF MEDICAL INFORMATICS},
 month = {SEP},
 researcherid-numbers = {Kim, Hyeoneui/NOE-3592-2025},
 title = {Evaluating the impact of explainable AI on clinicians' decision-making:
A study on ICU length of stay prediction},
 unique-id = {WOS:001488109100001},
 volume = {201},
 year = {2025}
}

@article{kayadibi2025ai,
 article-number = {105724},
 author = {Kayadibi, Ismail and Kose, Utku and Guraksin, Gur Emre and Cetin, Bilgun},
 doi = {10.1016/j.ijmedinf.2024.105724},
 earlyaccessdate = {DEC 2024},
 eissn = {1872-8243},
 issn = {1386-5056},
 journal = {INTERNATIONAL JOURNAL OF MEDICAL INFORMATICS},
 month = {MAR},
 orcid-numbers = {Kayadibi, Ismail/0000-0002-1949-8211
},
 researcherid-numbers = {ÇETİN, Bilgün/IZQ-2110-2023
Kose, Utku/C-8683-2009
Güraksın, Gür/ABI-3335-2020
Kayadibi, İsmail/JQW-3761-2023},
 title = {An AI-assisted explainable mTMCNN architecture for detection of
mandibular third molar presence from panoramic radiography},
 unique-id = {WOS:001372377600001},
 volume = {195},
 year = {2025}
}

@article{rainey2024operationalizing,
 article-number = {e0000560},
 author = {Rainey, Clare and Bond, Raymond and Mcconnell, Jonathan and Hughes,
Ciara and Kumar, Devinder and Mcfadden, Sonyia},
 doi = {10.1371/journal.pdig.0000560},
 eissn = {2767-3170},
 journal = {PLOS DIGITAL HEALTH},
 month = {AUG},
 number = {8},
 orcid-numbers = {Mc Fadden, Sonyia/0000-0002-4001-7769
Rainey, Clare/0000-0003-0449-8646},
 title = {Reporting radiographers' interaction with Artificial Intelligence-How do
different forms of AI feedback impact trust and decision switching?},
 unique-id = {WOS:001417669500001},
 volume = {3},
 year = {2024}
}

@article{singla2023medical,
 article-number = {102721},
 author = {Singla, Sumedha and Eslami, Motahhare and Pollack, Brian and Wallace,
Stephen and Batmanghelich, Kayhan},
 doi = {10.1016/j.media.2022.102721},
 earlyaccessdate = {DEC 2022},
 eissn = {1361-8423},
 issn = {1361-8415},
 journal = {MEDICAL IMAGE ANALYSIS},
 month = {FEB},
 orcid-numbers = {Singla, Sumedha/0000-0003-3477-0524
Batmanghelich, Kayhan/0000-0001-9893-9136},
 title = {Explaining the black-box smoothly-A counterfactual approach},
 unique-id = {WOS:000917473200001},
 volume = {84},
 year = {2023}
}

@incollection{shapley1953value,
  author    = {Shapley, Lloyd S.},
  title     = {A Value for n-person Games},
  booktitle = {Contributions to the Theory of Games, Vol. II},
  editor    = {Kuhn, H. W. and Tucker, A. W.},
  pages     = {307--317},
  publisher = {Princeton University Press},
  year      = {1953},
}

@ARTICLE{perez2015bigdata,
  author={Andreu-Perez, Javier and Poon, Carmen C. Y. and Merrifield, Robert D. and Wong, Stephen T. C. and Yang, Guang-Zhong},
  journal={IEEE Journal of Biomedical and Health Informatics}, 
  title={Big Data for Health}, 
  year={2015},
  volume={19},
  number={4},
  pages={1193-1208},
  doi={10.1109/JBHI.2015.2450362}}

@inproceedings{mothilal2020dice,
author = {Mothilal, Ramaravind K. and Sharma, Amit and Tan, Chenhao},
title = {Explaining machine learning classifiers through diverse counterfactual explanations},
year = {2020},
isbn = {9781450369367},
publisher = {Association for Computing Machinery},
address = {New York, NY, USA},
url = {https://doi.org/10.1145/3351095.3372850},
doi = {10.1145/3351095.3372850},
booktitle = {Proceedings of the 2020 Conference on Fairness, Accountability, and Transparency},
pages = {607–617},
numpages = {11},
location = {Barcelona, Spain},
series = {FAT* '20}
}

@misc{goldstein2013peeking,
  author = {Goldstein, Alex and Kapelner, Adam and Bleich, Justin and Pitkin, Emil},
  title = {Peeking Inside the Black Box: Visualizing Statistical Learning with Plots of Individual Conditional Expectation},
  howpublished = {arXiv:1309.6392},
  year = {2013},
  note = {Available at https://arxiv.org/abs/1309.6392}
}

@inproceedings{zhou2016learning,
  title     = {Learning Deep Features for Discriminative Localization},
  author    = {Zhou, Bolei and Khosla, Aditya and Lapedriza, Agata and Oliva, Aude and Torralba, Antonio},
  booktitle = {Proceedings of the IEEE Conference on Computer Vision and Pattern Recognition (CVPR)},
  pages     = {2921--2929},
  year      = {2016},
  doi       = {10.1109/CVPR.2016.319}
}

\appendix
\section{Keywords Selection} \label{sec:appendix_keywords}
Queries and explanations are available at this open GitHub repository: \url{https://github.com/iamalegambetti/CDSS-XAI-HCE-Survey}.

Our results are updated until October 2025.

\end{document}